\documentclass[letterpaper, 10 pt, conference]{ieeeconf}  

\IEEEoverridecommandlockouts                              

\usepackage{cite}
\usepackage{graphics} 
\usepackage{epsfig} 
\let\labelindent\relax
\usepackage[shortlabels]{enumitem}

\usepackage{mathptmx} 
\usepackage{times} 
\usepackage{amsmath} 
\usepackage{amssymb}  

\usepackage{color}
\usepackage[table]{xcolor}
\usepackage{epstopdf}
\DeclareGraphicsExtensions{.pdf,.png,.jpg}
\DeclareMathAlphabet{\pazocal}{OMS}{zplm}{m}{n}
 
\usepackage{multirow}

\usepackage[ruled]{algorithm2e}

\newcommand{\algfref}[1]{Algorithm~\ref{#1}}
\newcommand{\algref}[1]{Alg.~\ref{#1}}
\newcommand{\figfref}[1]{Figure~\ref{#1}}
\newcommand{\figref}[1]{Fig.~\ref{#1}}
\newcommand{\Figref}[1]{Figure~\ref{#1}}

\newcommand{\Tabref}[1]{Table~\ref{#1}}

\newcommand{\eqnref}[1]{Eq.~(\ref{#1})}
\newcommand{\secref}[1]{Sec.~\ref{#1}}

\newcommand{\etal}{\textit{et~al.}}

\title{\LARGE \bf
Camera Exposure Control for Robust Robot Vision \\ with Noise-Aware Image Quality Assessment
}

\author{Ukcheol Shin$^{1}$, Jinsun Park$^{1}$, Gyumin Shim$^{1}$, Francois Rameau$^{1}$ and In So Kweon$^{1}$
\thanks{$^{1}$The authors are with the Robotics and Computer Vision Laboratory, School of Electrical Engineering, KAIST, Daejeon, 34141, Republic of Korea. {\tt \textbraceleft shinwc159, zzangjinsun, shimgyumin, frameau, iskweon77\textbraceright @kaist.ac.kr}}%
}

\begin{document}

\maketitle
\thispagestyle{empty}
\pagestyle{empty}

\begin{abstract}

In this paper, we propose a noise-aware exposure control algorithm for robust robot vision. Our method aims to capture the best-exposed image which can boost the performance of various computer vision and robotics tasks. 
For this purpose, we carefully design an image quality metric which captures complementary quality attributes and ensures light-weight computation. Specifically, our metric consists of a combination of image gradient, entropy, and noise metrics.
The synergy of these measures allows preserving sharp edge and rich texture in the image while maintaining a low noise level.
Using this novel metric, we propose a real-time and fully automatic exposure and gain control technique based on the Nelder-Mead method.
To illustrate the effectiveness of our technique, a large set of experimental results demonstrates higher qualitative and quantitative performances when compared with conventional approaches.

\end{abstract}

\section{INTRODUCTION}
\label{sec:intro}

Capturing well-exposed images with low noise is crucial for computer vision and robotics algorithms. However, cameras suffer from fundamental hardware limitations such as a narrow dynamic range, small aperture size, and low sensor sensitivity, to name just a few. For example, a camera with a narrow dynamic range tends to acquire saturated images under challenging conditions. Moreover, motion blur or severe noise can occur in low-light environments due to long exposure time or large gain, respectively. Although we cannot fully avoid these limitations, they can be drastically alleviated by carefully adjusting camera exposure parameters such as the aperture size, exposure time, and gain.

Prior to camera exposure parameter control, a well-defined image quality metric is required. To be effective, robust, and generic, this metric has to incorporate various criteria such as the average image brightness, sharpness, noise, and saturation. However, most approaches rely on camera built-in Auto-Exposure (AE) algorithms~\cite{muramatsu1997photometry, johnson1984photographic, sampat1999system} or use a fixed exposure time manually set by the user. These approaches do not guarantee an optimal quality image and lead to severe image degradation. 


Based on this observation, several approaches~\cite{torres2015optimal,lu2010camera,shim2014auto,zhang2017active,kim2018exposure} reveal that utilizing more meaningful and appropriate image statistics such as gradient distribution and entropy is essential to improve the camera exposure control. Although the image gradient- or entropy-based metrics demonstrate satisfying results in various environments, these techniques do not consider the image noise in their estimation. This omission leads to high gain causing strong salt-and-pepper noise in the image, which is particularly disadvantageous for most robotics and computer vision tasks.



In this paper, we propose a novel image quality metric fusing low-level measurements and noise estimation. In addition, we propose a real-time exposure control algorithm based on the Nelder-Mead (NM) method~\cite{nelder1965simplex}. The proposed control algorithm ensures an efficient searching strategy and converges to the best exposure parameters according to the proposed metric. A large set of experiments including feature matching, pose estimation, object detection, and computational cost analysis emphasizes the superiority and effectiveness of the proposed algorithm.

\begin{figure}[t]
\begin{center}
\footnotesize
\begin{tabular}{@{}c@{\hskip 0.01\linewidth}c}
\includegraphics[width=0.33\linewidth]{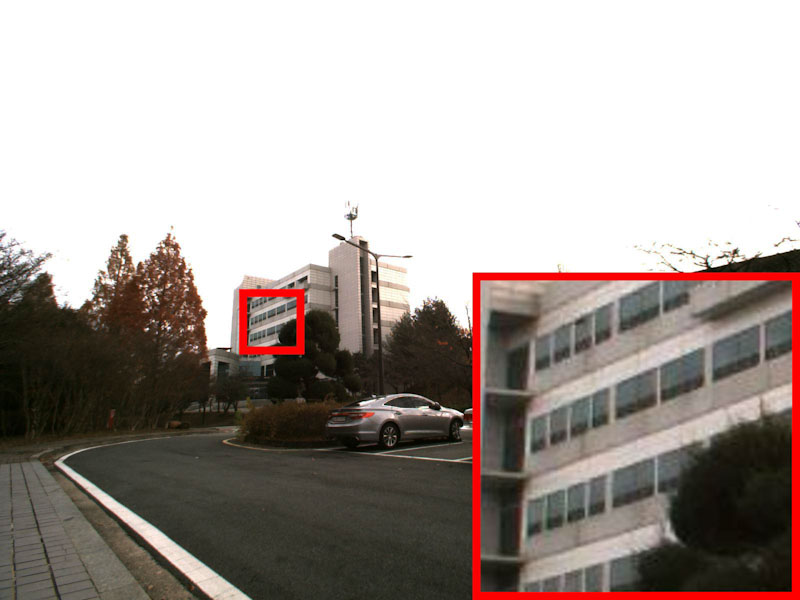} &
\includegraphics[width=0.33\linewidth]{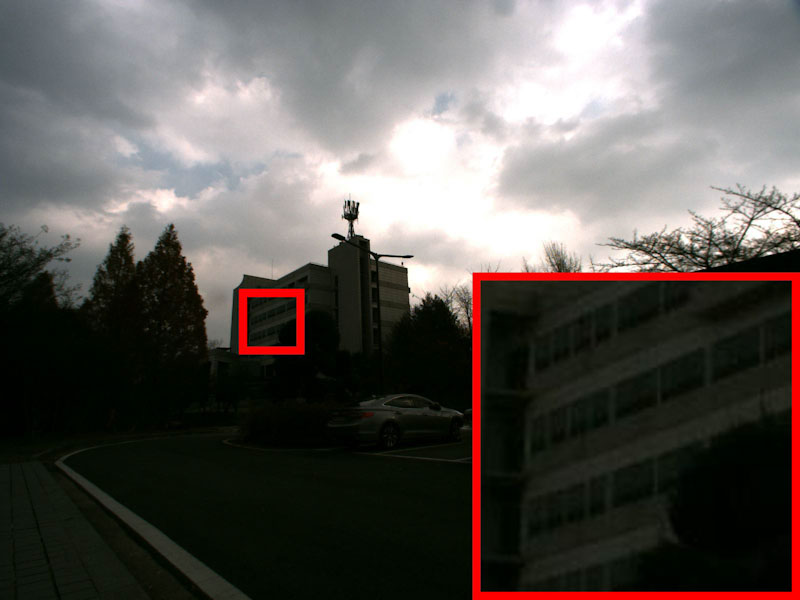} \\
Ours $(2dB, 6.6ms)$ & AE $(0dB, 0.7ms)$ \\
\end{tabular}
\begin{tabular}{@{}c@{\hskip 0.01\linewidth}c@{\hskip 0.01\linewidth}c}
\includegraphics[width=0.33\linewidth]{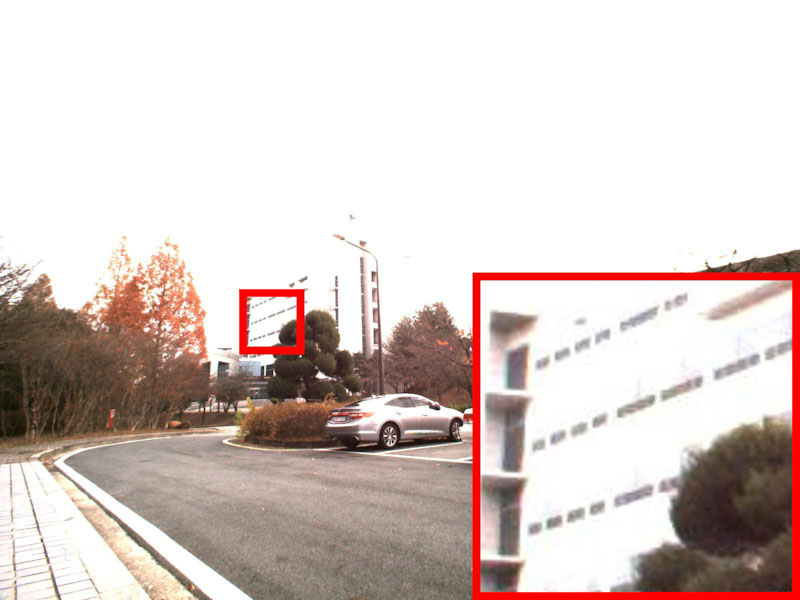} &
\includegraphics[width=0.33\linewidth]{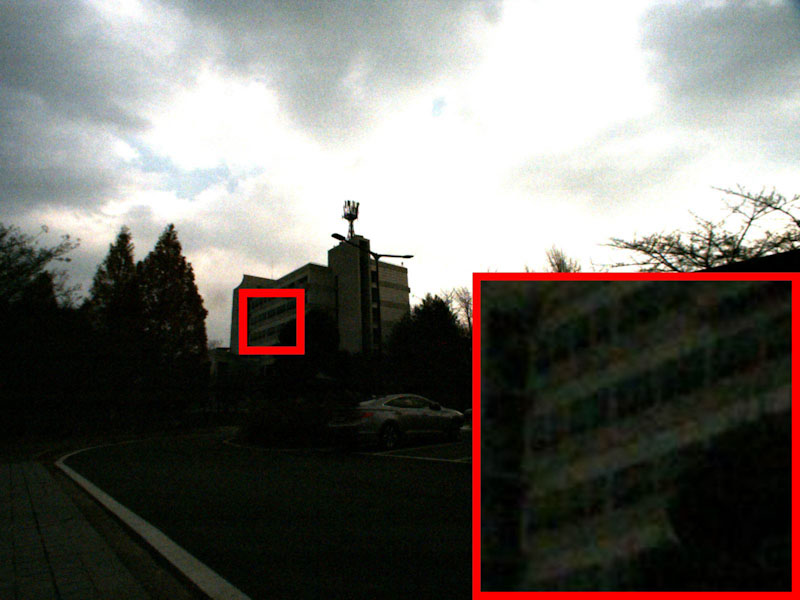} &
\includegraphics[width=0.33\linewidth]{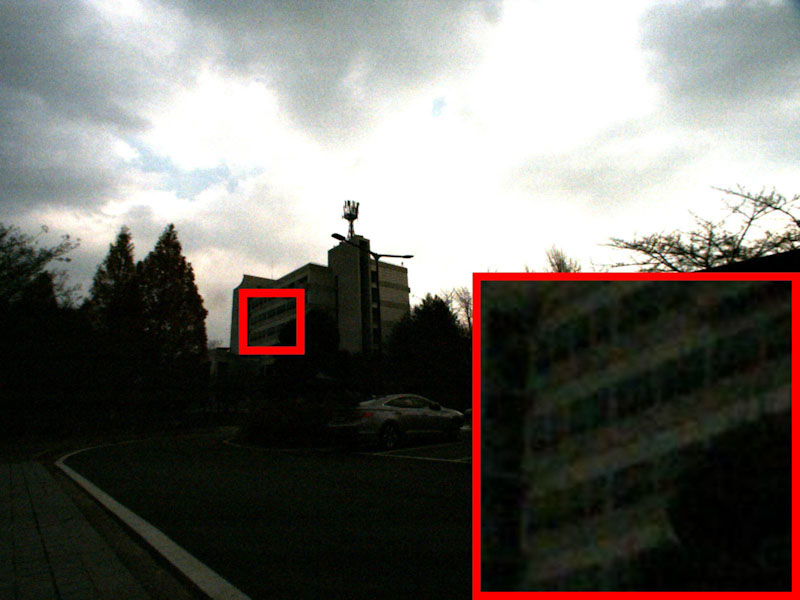} \\
Shim~\cite{shim2014auto} $(10dB, 6.9ms)$ & Zhang~\cite{zhang2017active} $(20dB, 0.3ms)$ & Kim~\cite{kim2018exposure} $(20dB, 0.3ms)$
\end{tabular}
\end{center}
\vspace{-0.1in}
\caption{{\bf Images captured by each AE algorithm under same environment with (gain($dB$), exposure time($ms$))}. Our algorithm selects high quality image with less noise successfully compared to the other approaches.}
\label{fig:teaser}
\vspace{-0.2in}
\end{figure}

\begin{figure*}[t]
\begin{center}
 \includegraphics[width=0.90\linewidth]{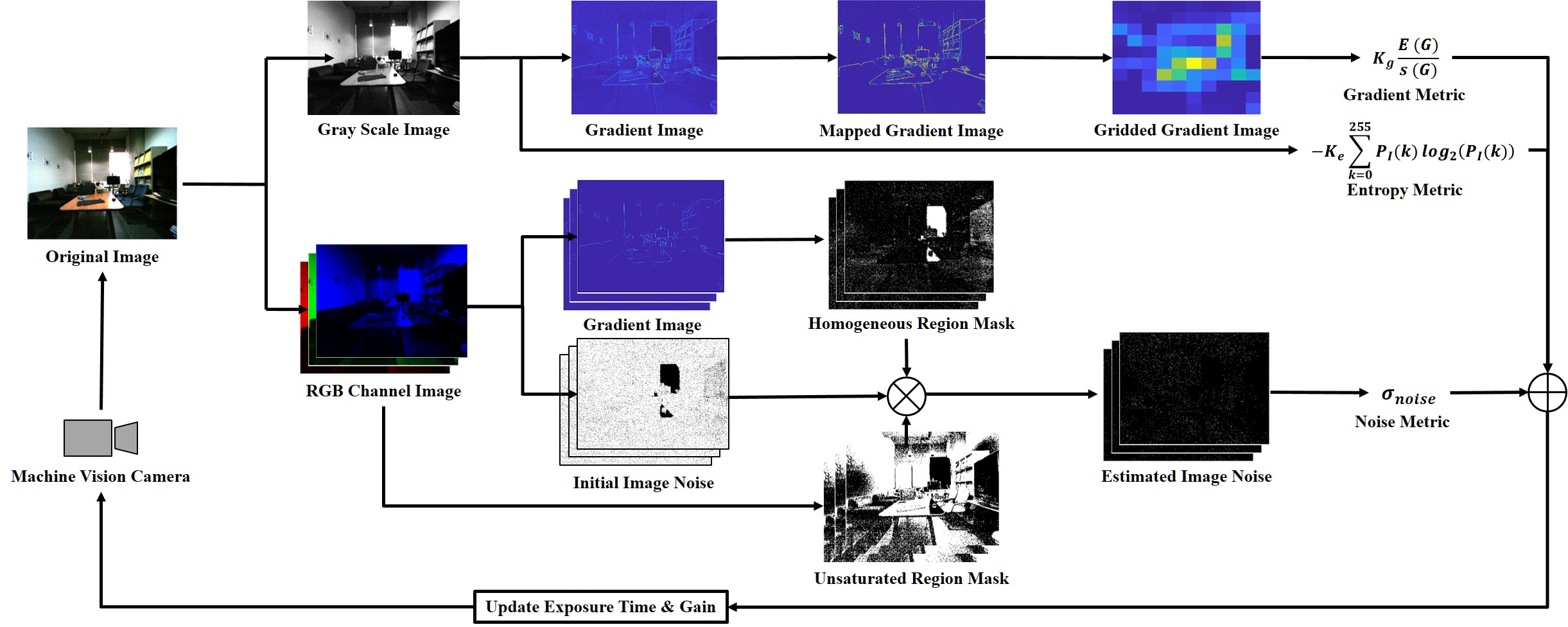}
\end{center}
\vspace{-0.2in}
\caption{\textbf{Overall pipeline of the proposed algorithm.} The proposed algorithm measures image quality based on three image properties: image gradient, entropy, and noise. After that, we update exposure time and gain using the Nelder-Mead method until we get a well-exposed image.}
\label{fig:pipe_line}
\vspace{-0.2in}
\end{figure*}

\section{Related Work}
\label{sec:relatedwork}

Capturing a well-exposed image is an essential condition to apply any vision based algorithms under challenging environments. In this paper, we define the term \textit{`well-exposed'} from a robotics point of view, as an image containing texture details, sharp object boundaries with low noise, saturation, and blur. In fact, these conditions are desirable to achieve various tasks such as visual-SLAM~\cite{mur2017orb} that requires robust and repeatable keypoints detection, instance segmentation~\cite{he2017mask} that requires sharp object boundaries, and object classification where even an imperceptible noise may lead to misclassification~\cite{moosavi2017universal}.

To capture a well-exposed image, a criterion that quantifies the quality of an image is required. Once a reliable metric is established, it is possible to dynamically adjust the camera exposure parameters such that they maximize the quality criterion. Many attempts focus on the definition of a reliable metric for auto-exposure control, one of the most commonly used metric is the image intensity histogram~\cite{torres2015optimal,neves2009autonomous}. Although these crude approaches are very fast, they admit a low robustness against illumination changes and complex scenarios.


Alternatively, other approaches rely on image gradients to maximize the quantity of information in the image. For instance, Shim~\etal~\cite{shim2014auto} suggested a mapping function between the gradient magnitude and the gradient information. Based on this relation, the image with the largest gradient information is adopted as a well-exposed image. For the convergence of the algorithm, the authors proposed to control the exposure parameters via synthetic images generated by gamma correction. More recently, Zhang~\etal~\cite{zhang2017active} proposed another gradient-based metric where a weighted sum of sorted gradient magnitude allows finding the optimal exposure via a gradient descent algorithm. 

The main problem of gradient-based metrics is their tendency to favor high exposures, which, in turn leads to over-exposed images. To avoid such problem, Kim~\etal~\cite{kim2018exposure} proposed a gradient weighting scheme based on local image entropy. The optimal exposure is estimated via a Bayesian optimization framework, which finds the global solution by estimating the surrogate models. However, the complexity of the Bayesian optimization and weighting scheme does not allow real-time ability.

The limitations of gradient-based approaches are that they consider the exposure time only. Moreover, this kind of metric is also particularly sensitive to noise due to inappropriate gain. In order to tackle this issue, we propose a fast noise-aware image quality metric based on the quantity of image information (gradient, entropy) and the level of noise. Based on this proposed metric, our algorithm is able to obtain a well-exposed image with low noise effectively, as shown in~\figref{fig:teaser}. We also propose a NM method based real-time control algorithm that ensures fast and reliable convergence to optimal exposure time and gain simultaneously.

\section{Image Quality Metric and Exposure Parameter Control}
\label{sec:metricandcontrol}

The proposed algorithm consists of two main modules: the image quality assessment module and the exposure parameter control module. An illustration of our strategy is available in~\figref{fig:pipe_line}. We assess the image quality based on three image properties: image gradient, entropy, and noise. First, we compute grid-level statistics of the gradient and global entropy of an input image. Simultaneously, the image noise is estimated by inspecting unsaturated homogeneous regions in the input image. Based on the calculated image quality, the camera exposure parameters (\textit{i.e.}, exposure time and gain) are updated accordingly.



\subsection{Gradient-Based Metric}
\label{subsec:gradientmetric}

The purpose of the gradient-based metric is to effectively evaluate the texture and edge information contained in the image. For this purpose, we first adopt a mapping function proposed by Shim~\etal~\cite{shim2014auto}, then we further improve the performance by complementing its limitation. The mapping function is defined as follows:

\begin{multline} 
\label{equ:mapping_function}
\tilde{g_{i}} = 
\left\{\begin{array}{ccc}
\frac{1}{N_{g}}\log\left(\lambda\left(g_{i} - \gamma\right) + 1\right), & 
\textrm{for} & \textrm{$g_{i} \geq \gamma$} \\
\textrm{0}, & 
\textrm{for} & \textrm{$g_{i} < \gamma$} 
\end{array} 
\right. \\
\qquad \textrm{s.t. } N_{g} = \log\left(\lambda\left(1 - \gamma\right) + 1\right),
\end{multline}
where $g_{i} \in [0, 1]$\footnote{We assume that the pixel range is $[0, 255]$ and the gradient magnitude range is $[0, 1]$ due to implicit normalization factor $1/255$.} denotes the gradient magnitude at pixel $i$, $\gamma$ indicates the activation threshold value, $\lambda$ is the control parameter to adjust the mapping behavior, $N_{g}$ is the normalization factor, and $\tilde{g_{i}}$ stands for the amount of gradient information at pixel $i$. The interested reader may refer to \cite{shim2014auto} for further details.



The proposed mapping function eliminates meaningless gradient caused by image noise and adjusts the difference between strong and weak gradients. Therefore, it extracts useful gradient information, however, this function still favors strongly biased gradient in particular area caused by high exposure values. As a result, the details of the entire image are ignored. To resolve this problem, we additionally consider the uniformity of the gradient information. Our gradient-based image quality metric based on grid-level $\tilde{g_{i}}$ statistics is defined as follows:

\begin{gather}
G_{j} = \sum_{i \in C_{j}}{\tilde{g}_{i}},\quad j = 1, 2, \dots, N_{C}, \label{eqn:gridgradinfo} \\
L_{gradient} = K_{g} \cdot E(G) / s(G), \label{equ:loss_gradient}
\end{gather}
where $G_{j}$ is the $j$-th grid cell, $N_{C}$ denotes the total number of grid cells, $K_{g}$ is the normalization factor, $E(\cdot)$ and $s(\cdot)$ denote mean and standard deviation operators, respectively. 

We divide the mapped gradient image into $N_{C}$ grid cells, then aggregate the gradient information for each grid cell to measure the strength and uniformity of the gradient information. $E(G)$ and $s(G)$ represent the overall amount and the degree of dispersion of the information throughout the image, respectively. If $L_{gradient}$ is large, it means that the gradient information is strong and uniformly distributed. Otherwise, the gradient information is weak and biased.



\subsection{Entropy-Based Metric}
\label{subsec:entropymetric}

Although it is possible to grasp some image characteristics through the gradient-based metric, it cannot fully catch basic image attributes such as color, contrast, and brightness. To compensate this problem, we adopt the global image entropy which represents the amount of information contained in the image. By utilizing the global image entropy for our image quality metric, we are able to fully evaluate the quantity of the low-level image information. Our entropy-based image quality metric is defined as follows:

\begin{equation} 
\label{equ:entropy}
L_{entropy} = -K_{e}\sum_{k=0}^{255}{P_{I}(k)}\log_{2}{P_{I}(k)},
\end{equation}
where $P_{I}(k)$ denotes the probability of pixel value $k$ in the gray scale image and $K_{e}$ is the normalization factor.

\subsection{Noise-Based Metric}
\label{subsec:noisemetric}

Due to the noise induced by the gain, using the exclusive two metrics presented above is insufficient to ensure the capture of high quality images. Therefore, we take the image noise into consideration for our image quality assessment. For the image noise level estimation, eigenvalue analysis~\cite{chen2015efficient} gives highly accurate results, however, its computational cost is too high and inappropriate for real-time applications. Although filter-based approaches~\cite{immerkaer1996fast, yang2010fast} are less accurate compared to the eigenvalue analysis, they still give reliable noise estimation results with fast computation. Specifically, we construct our noise-based metric based on the filter-based approach~\cite{immerkaer1996fast}.

For the image noise estimation, we assume that the image noise is an additive zero-mean Gaussian noise. For this type of noise, Immerkaer~\cite{immerkaer1996fast} proposed the noise estimation kernel $M$ as follows:

\begin{equation} 
\label{equ:noisefilter}
M = {\footnotesize \begin{bmatrix} 1 & -2 & 1 \\ -2 & 4 & -2 \\ 1 & -2 & 1 \end{bmatrix}}.
\end{equation}

Using the noise estimation kernel $M$, we are able to estimate the noise level of the entire image. However, the noise level estimated on the whole image is inaccurate, because the noise estimation kernel $M$ is sensitive to object structures. Therefore, noise estimation kernel $M$ should be applied only on the homogeneous regions for better accuracy. For this purpose, we define the homogeneous region mask $H$ as follows:



\begin{equation} 
\label{equ:homogeneousmask}
H(i) = 
\left\{\begin{array}{ccc}
1, & \textrm{for} & g_{i} \leq \delta \\
0, & \textrm{for} & g_{i} > \delta
\end{array},
\right.
\end{equation}
where the adaptive threshold $\delta$ is the $p$-th percentile of gradients in the image. From the mask $H$, we can effectively extract homogeneous regions. However, under-/over-saturated regions must be excluded from $H$ since they cannot contain noise due to the saturation. Therefore, the unsaturated region mask $U$ is defined based on a simple threshold scheme as follows:


\begin{equation} 
\label{equ:unsaturationmask}
U(i) = 
\left\{\begin{array}{ccc}
1, & \textrm{for} & \tau_{l} \leq I(i) \leq \tau_{h} \\
0, & \textrm{for} & \textrm{otherwise}
\end{array},
\right.
\end{equation}
where $\tau_{l}$ and $\tau_{h}$ denotes lower and upper bounds for unsaturated pixel values, respectively. After we obtain unsaturated homogeneous regions from \eqnref{equ:homogeneousmask} and \eqnref{equ:unsaturationmask}, we estimate the noise variance of the image as follows:


\begin{equation} 
\label{equ:our sigma}
\sigma_{noise} = \sqrt[]{\frac{\pi}{2}}\frac{1}{N_{S}}\sum_{i} H(i) \cdot U(i) \cdot |I*M|(i),
\end{equation}
where $N_{S}$ denotes the number of valid pixels in the mask $H \cdot U$, $*$ denotes the convolution operator and $|\cdot|$ denotes the absolute operator. We utilize $\sigma_{noise}$ as our noise-based image quality metric. For color images, we estimate the noise levels in each color channel and the estimated values are averaged.

\subsection{Image Quality Metric}
\label{subsec:qualitymetric}
By combining gradient-, entropy- and noise-based metrics, our image quality metric is defined as follows:

\begin{equation} 
f(I) = \alpha \cdot L_{gradient} + (1-\alpha) \cdot L_{entropy} - \beta \cdot \sigma_{noise},
\label{equ:ourmetric}
\end{equation}
where $\alpha$ and $\beta$ are user parameters to adjust the effect of each term. A high $f(I)$ value implies that the image $I$ has low noise level and saturation with abundant texture details, which is the desired output of our algorithm. Therefore, we try to maximize $f(I)$ to obtain a well-exposed image by controlling camera exposure parameters.

\begin{figure*}[th]
\centering
\begin{tabular}
{@{}c@{\hskip 0.005\linewidth}c@{\hskip 0.005\linewidth}c@{\hskip 0.005\linewidth}c@{\hskip 0.005\linewidth}c}
\includegraphics[width=0.195\linewidth]{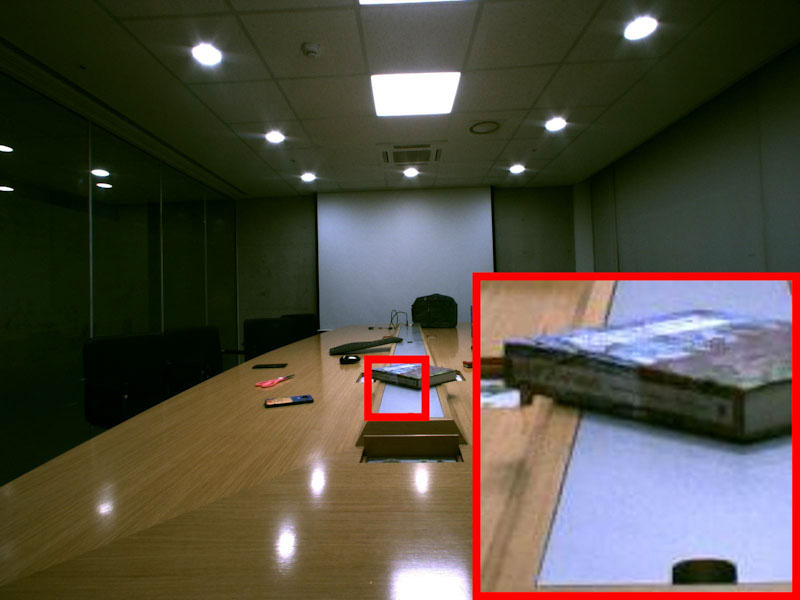} &
\includegraphics[width=0.195\linewidth]{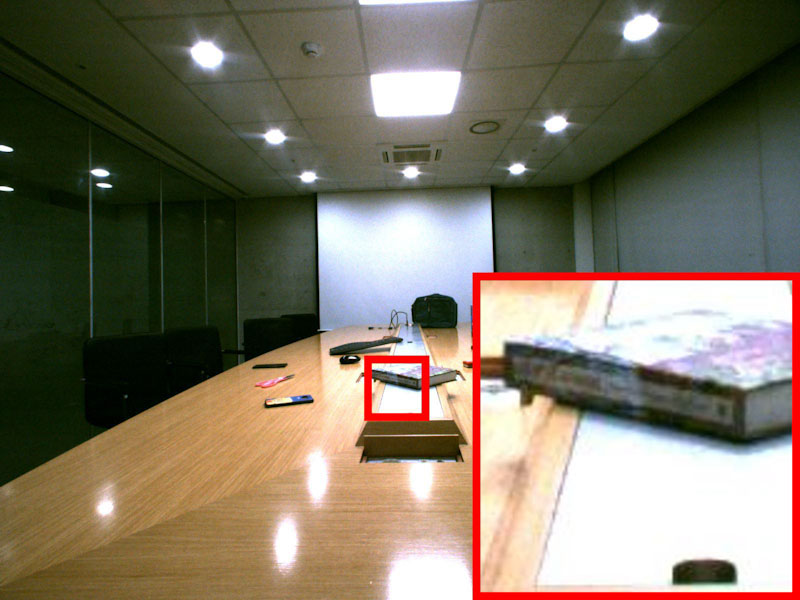} &
\includegraphics[width=0.195\linewidth]{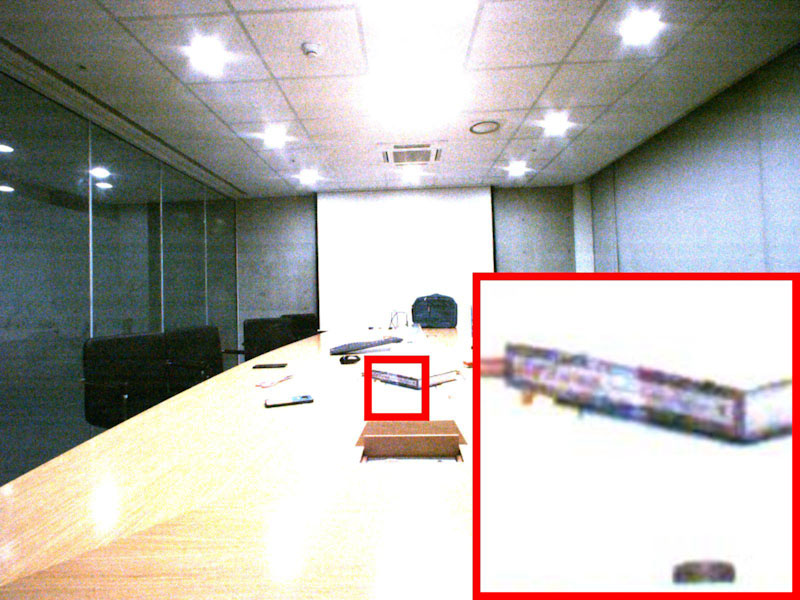} &
\includegraphics[width=0.195\linewidth]{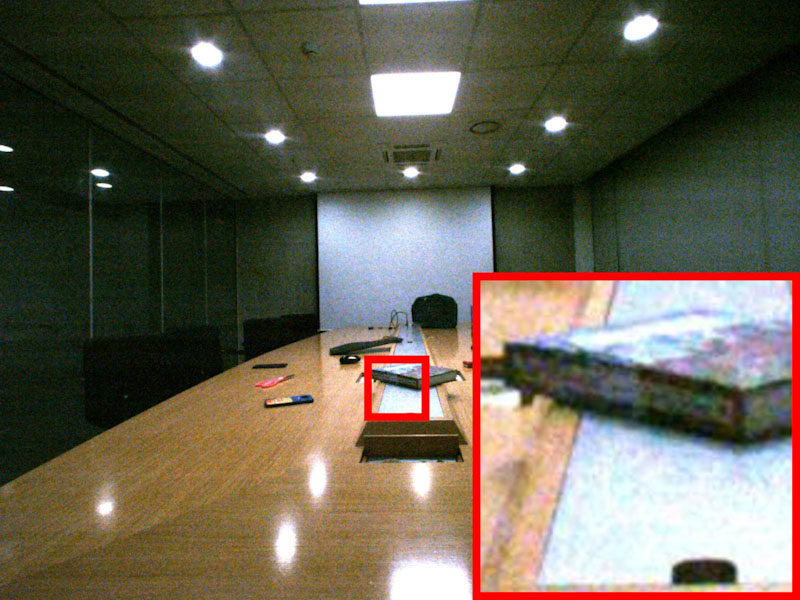} &
\includegraphics[width=0.195\linewidth]{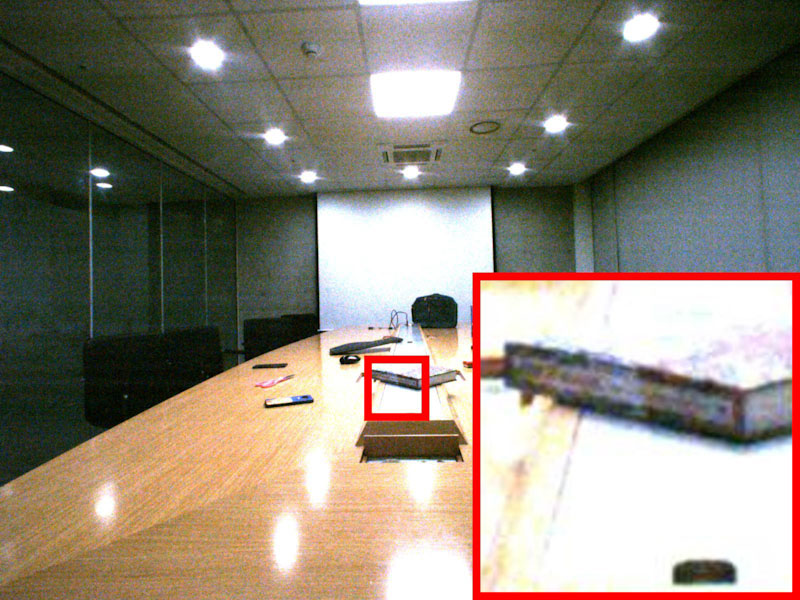} \\
{\footnotesize \textit{Ours} $(1dB, 67ms)$} & {\footnotesize \textit{AE} $(12.5dB, 30ms)$} & {\footnotesize \textit{Shim} $(24dB, 16ms)$} & {\footnotesize \textit{Zhang} $(24dB, 7ms)$} & {\footnotesize \textit{Kim} $(24dB, 10ms)$} \\
\includegraphics[width=0.195\linewidth]{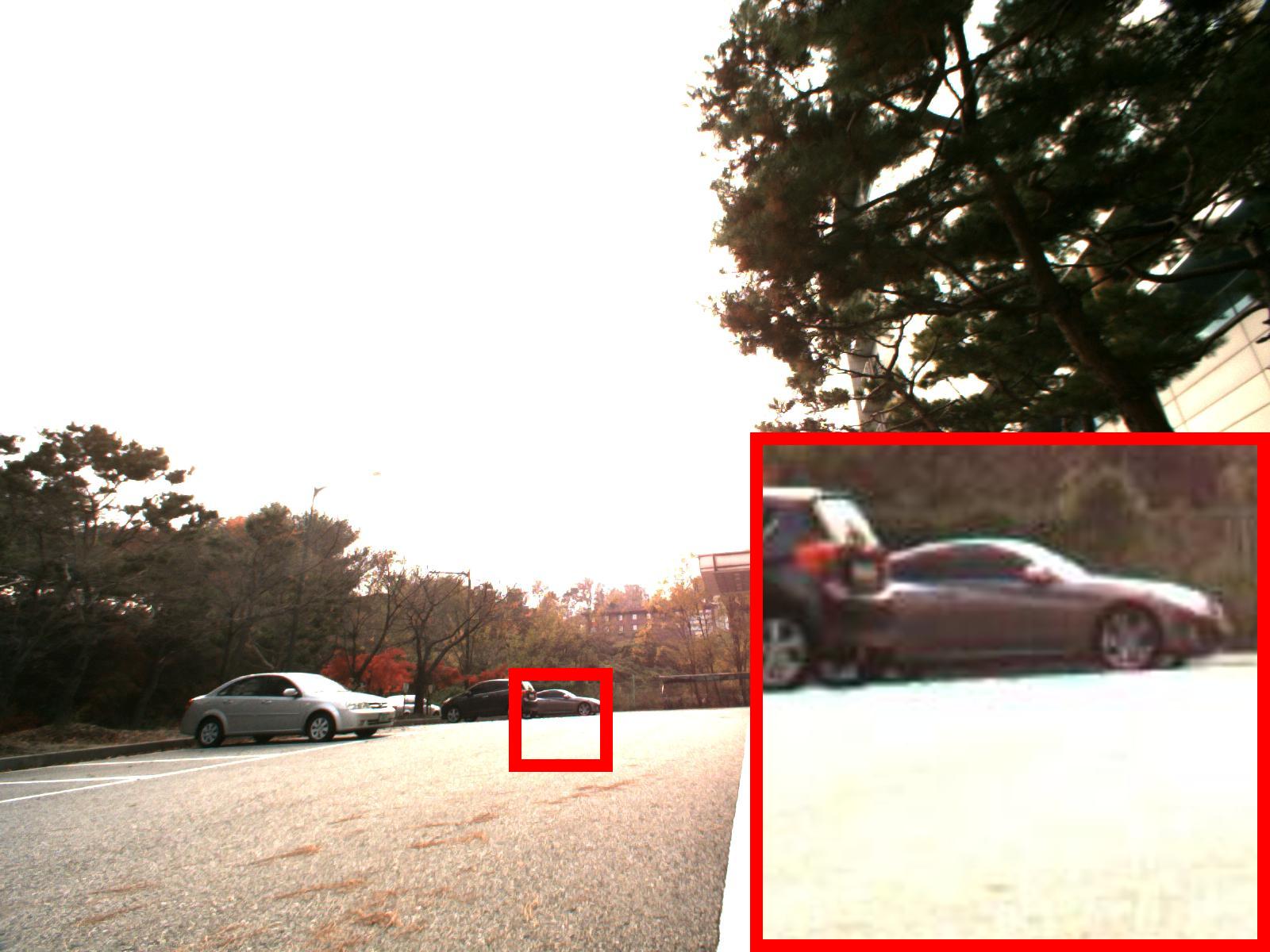} &
\includegraphics[width=0.195\linewidth]{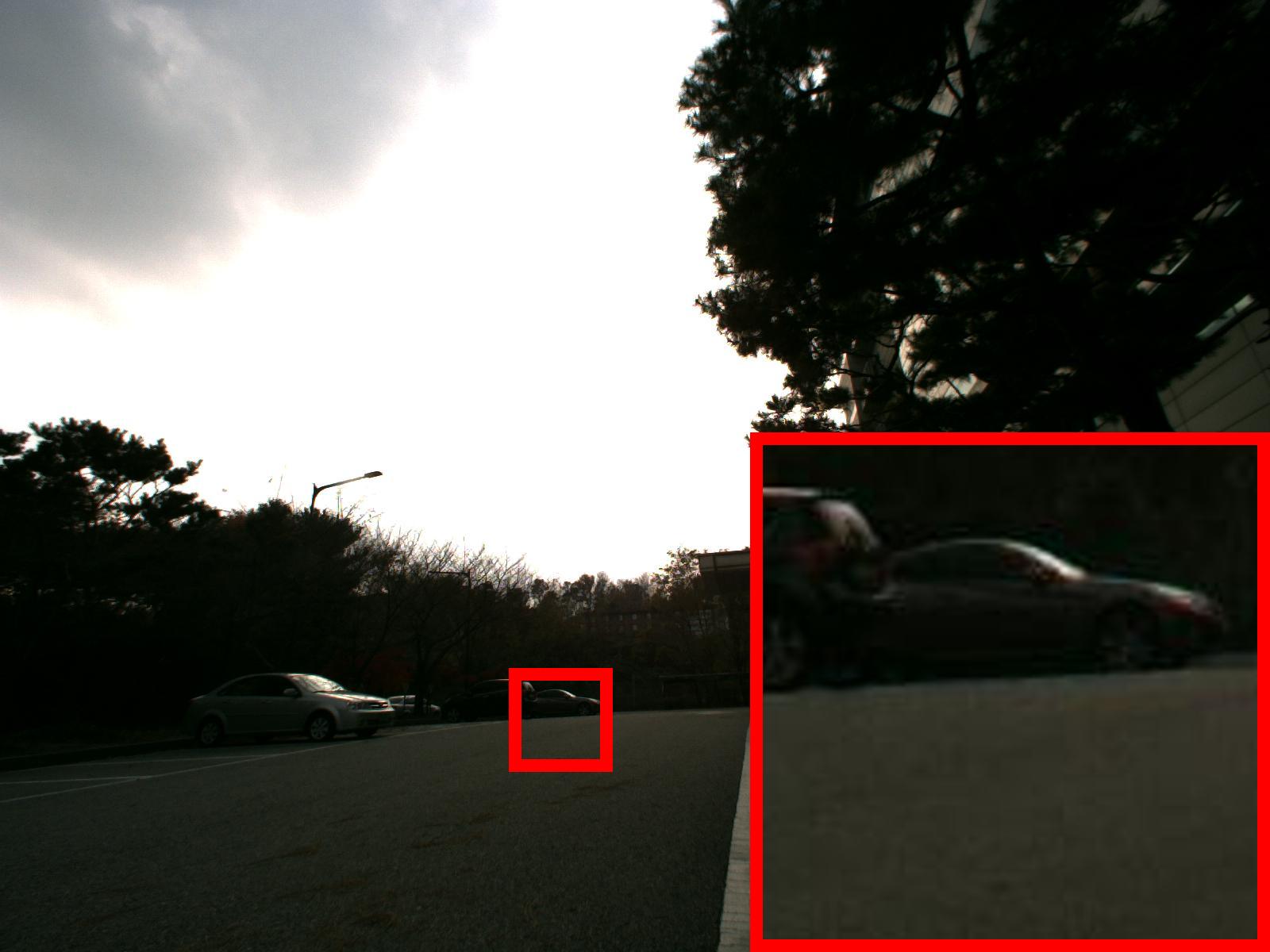} &
\includegraphics[width=0.195\linewidth]{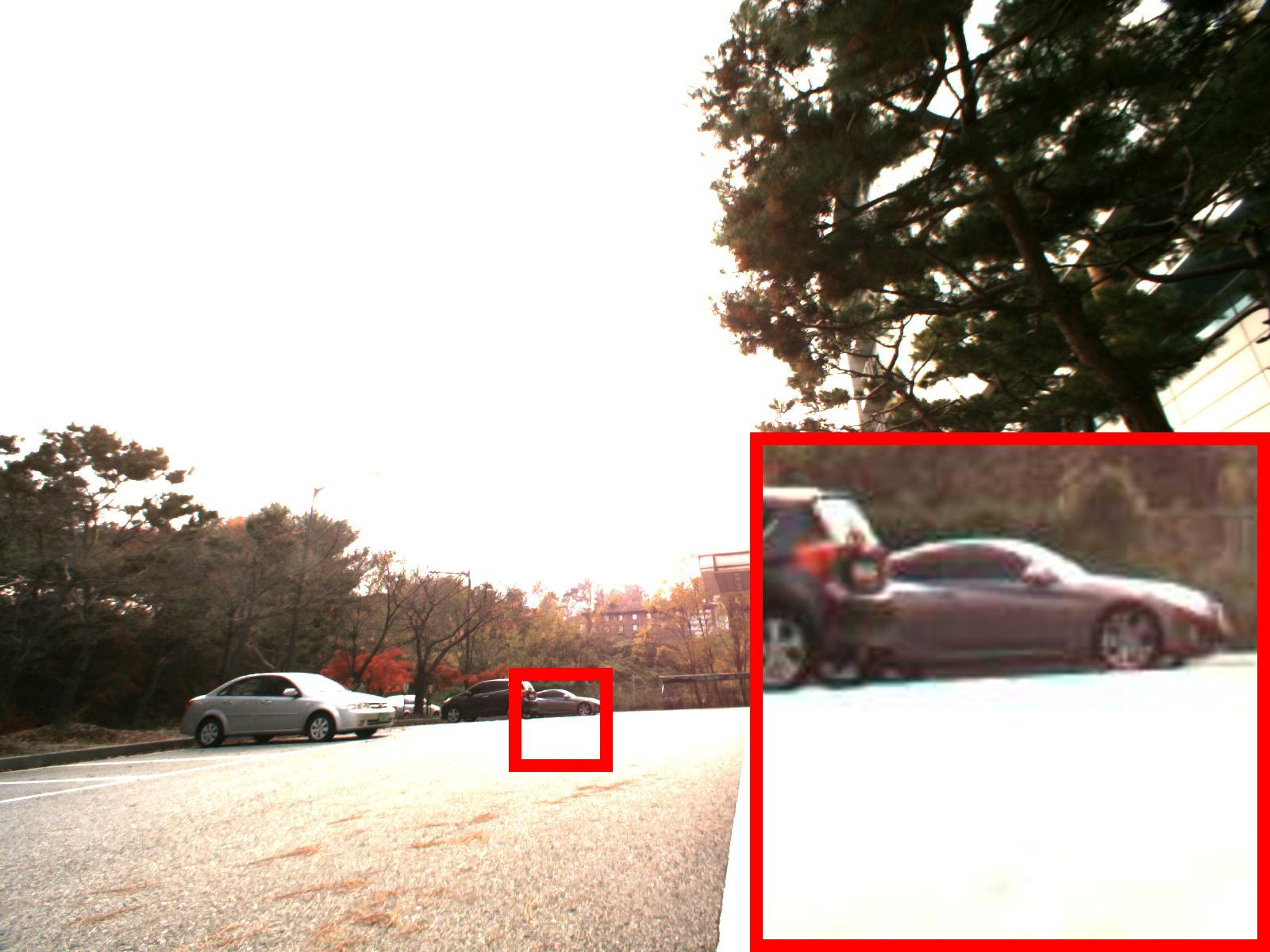} &
\includegraphics[width=0.195\linewidth]{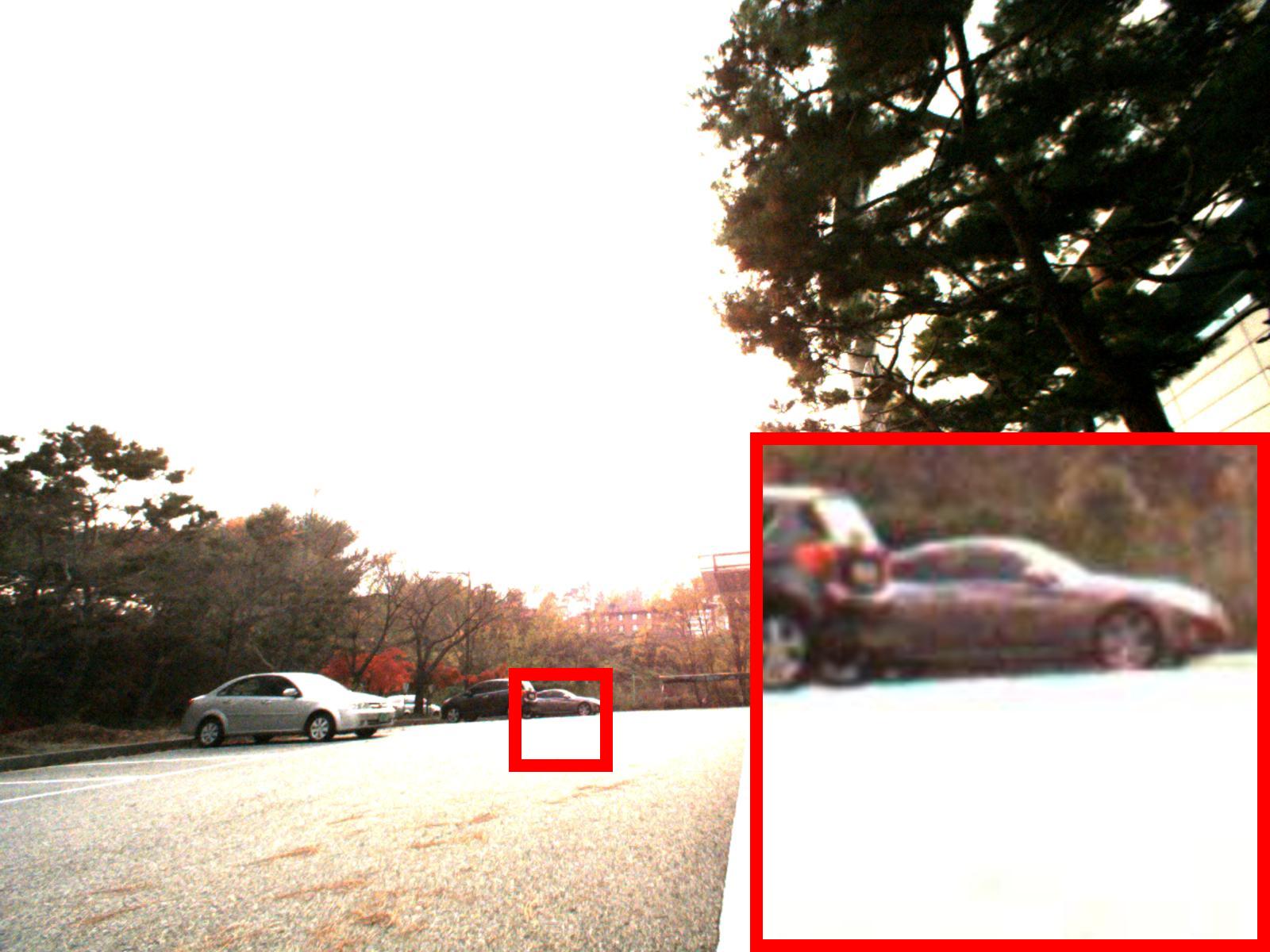} &
\includegraphics[width=0.195\linewidth]{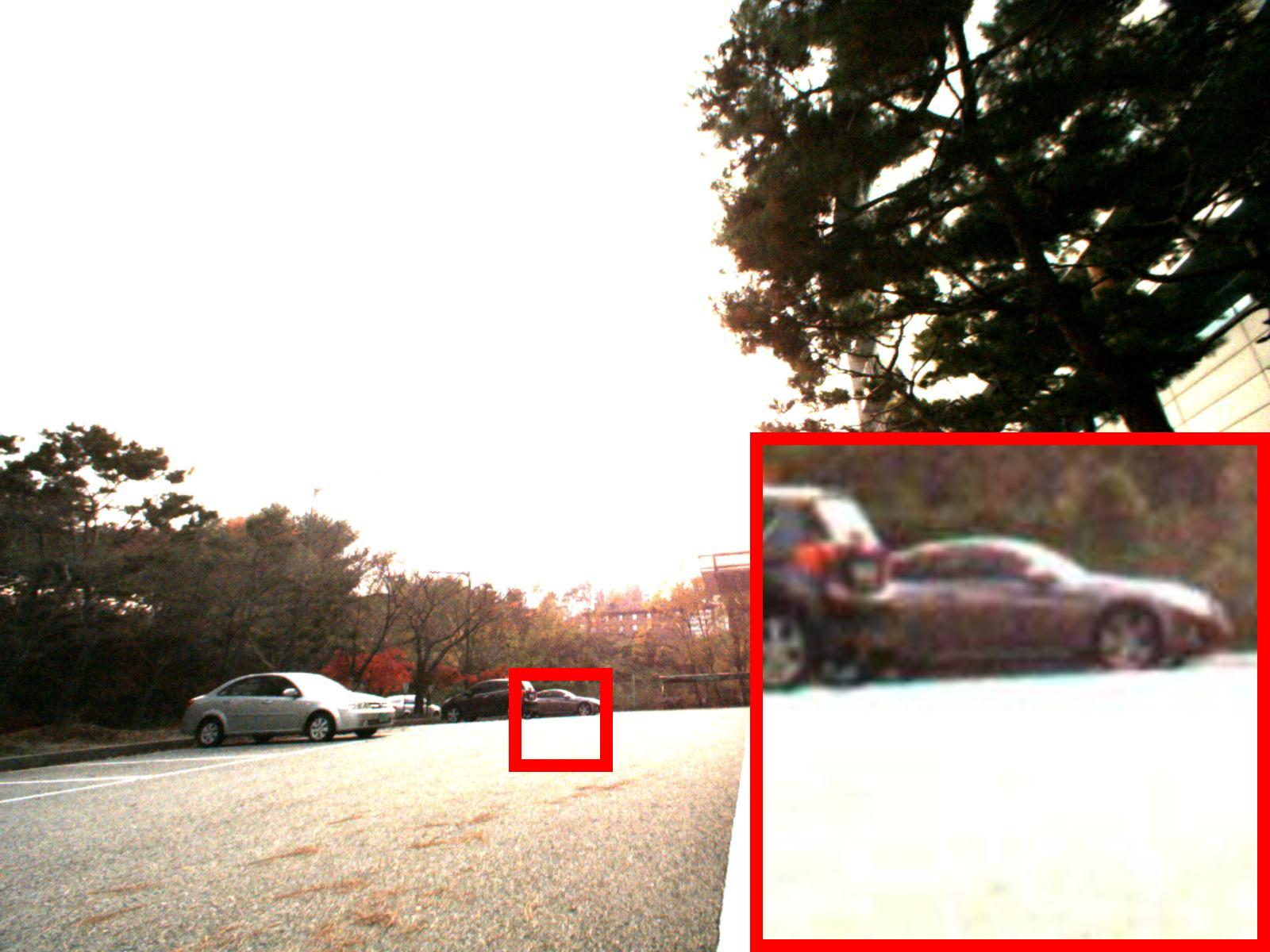} \\
{\footnotesize \textit{Ours} $(4dB, 6.7ms)$} & {\footnotesize \textit{AE} $(0dB, 2.3ms)$} & {\footnotesize \textit{Shim} $(4dB, 7.5ms)$} & {\footnotesize \textit{Zhang} $(20dB, 1.6ms$)} & {\footnotesize \textit{Kim} $(20dB, 1.5ms)$} \\
\end{tabular}
\vspace{-0.05in}
\caption{{\bf Illustrations of optimal images captured by each AE algorithms with (gain($dB$), exposure time($ms$)).} Our algorithm selects high quality images with less noise successfully compared to the other approaches. Highlighted regions show the noise and details in each image.}
\vspace{-0.2in}
\label{fig:noise_illustration}
\end{figure*}

\setlength{\textfloatsep}{0pt}
\setlength{\intextsep}{0pt} 
\begin{algorithm}[h]
    \label{alg:NMalgorithm}
    \caption{NM based Camera Exposure Control}
    \scalebox{0.95}{%
    \vbox{%
      \KwIn{Current image and exposure parameters $\mathbf{x}_{0}=\left[ExpT_0, Gain_0\right]$}
      \begin{enumerate}
        \item
        Construct initial simplex :
        \begin{enumerate}[(a)]
          \item
                Compute mean intensity $J_{\mathbf{x}_{0}}$ of the current image
          \item
                Compute step size $h$ in the direction of unit vector $\mathbf{e}_{i} \in \mathbb{R}^{2}$ where $i \in \{1, 2\}$
            \begin{equation*} 
            h = 
            \left\{\begin{array}{ccc}
            -\epsilon^{-1} (J_{\mathbf{x}_{0}}/255), & \textrm{for} & 128 \leq J_{\mathbf{x}_{0}} \leq 255  \\
            \epsilon(1 - J_{\mathbf{x}_{0}}/255), & \textrm{for} & 0 \leq J_{\mathbf{x}_{0}} < 128
            \end{array}
            \right.
            \end{equation*}
            \vspace{-0.1in}        
          \item
            Compute vertices of initial simplex.
            \begin{equation*} 
            \mathbf{x}_{i} = \mathbf{x}_{0}\cdot(1 + h\mathbf{e}_{i}), \qquad i \in \{1, 2\}    
            \end{equation*}
            \vspace{-0.2in}        
        \end{enumerate}
    
        \item
        Update the simplex : 
        \begin{enumerate}[(a)]
             \item
              Order according to the evaluations through ~\eqnref{equ:ourmetric} at the vertices of the simplex and decide the \textit{worst, second worst,} and the \textit{best} vertices.
             \item
              Calculate the centroid $\mathbf{x}_c$ of all points except for the $\mathbf{x}_{worst}$.
             \item
              Update simplex using \textit{reflection, expansion, contraction,} or \textit{shrink} operations with the objective function~\eqnref{equ:ourmetric}.
              \item
              Repeat from step (a) until the stopping criteria is satisfied.
            \end{enumerate}
      \item
      Return the output $\mathbf{x}_{opt} = \mathbf{x}_{best} = \left[ExpT_{opt} , Gain_{opt}\right]$.
      \end{enumerate}
    }}
\end{algorithm}

\subsection{Camera Exposure Parameter Control}
\label{subsec:paramcontrol}

To solve the maximization problem of $f(I)$, we utilize the Nelder-Mead (NM) method~\cite{nelder1965simplex,lagarias1998convergence} which ensures an efficient searching strategy and real-time performance. It does not require any derivative of object function by using the concept of a simplex, which is a special polytope of $n+1$ vertices in $n$ dimensions. The main problem in applying the NM method to the exposure parameter control problem is to provide an appropriate initial simplex; A small simplex leads to local maxima while a large one can cause drastic variation of the image. To resolve this problem, we have designed an efficient initial simplex construction method suitable for the exposure control problem. Note that the NM method does not ensure the global solution, but the proper initialization makes the convergence reliable. 


The designed initialization method decides the proper initial vertices of the simplex according to the mean intensity of the given image. Thereafter, the objective function~\eqnref{equ:ourmetric} is maximized as the operation of the NM method proceeds, then the optimal exposure parameters are obtained. \algfref{alg:NMalgorithm} describes our noise-aware exposure control algorithm to maximize the objective function~\eqnref{equ:ourmetric}. In this algorithm, $\mathbf{x}$ denotes the camera exposure parameters consists of exposure time ($ExpT$) and gain. While, $\epsilon$ is the scaling factor for the step size $h$.

\section{Experiments}
\label{sec:experiments}

In this section, we first describe our exposure control dataset. Afterward, various experiments including feature matching, pose estimation, object detection, and noise estimation are presented to demonstrate the performance of the proposed algorithm. We also conduct an ablation study to verify the role of individual metrics in the image quality assessment. In addition, we analyse the convergence speed and reliability of the proposed control scheme. Lastly, we analyze the processing time and show the real-time ability of the proposed algorithm. Throughout experiments, we set $\gamma = 0.06, \lambda = 10^{3}, N_{C} = 100, p = 0.1, \tau_{l} = 15, \tau_{h} = 235, K_{g} = 2, K_{e} = 0.125$, $\alpha = \beta = 0.4$, and $\epsilon$ =1.7.


\begin{figure*}[t]
\centering
\footnotesize
\begin{tabular}
{@{}c@{\hskip 0.001\linewidth}c@{\hskip 0.001\linewidth}c@{\hskip 0.001\linewidth}c@{\hskip 0.001\linewidth}c}
\includegraphics[width=0.195\linewidth]{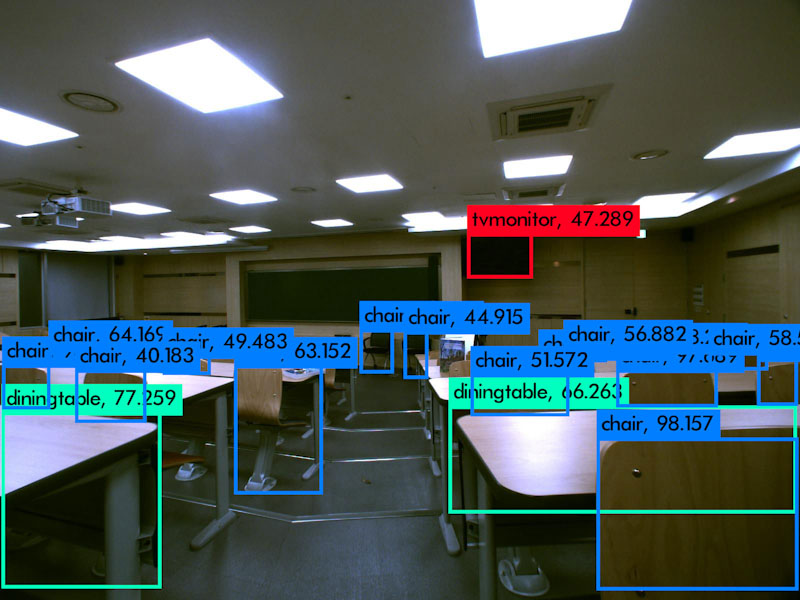} &
\includegraphics[width=0.195\linewidth]{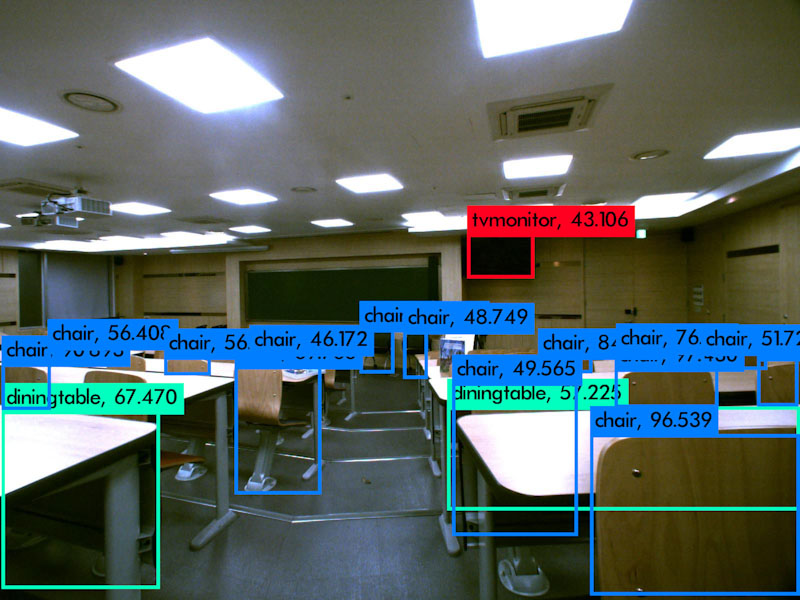} &
\includegraphics[width=0.195\linewidth]{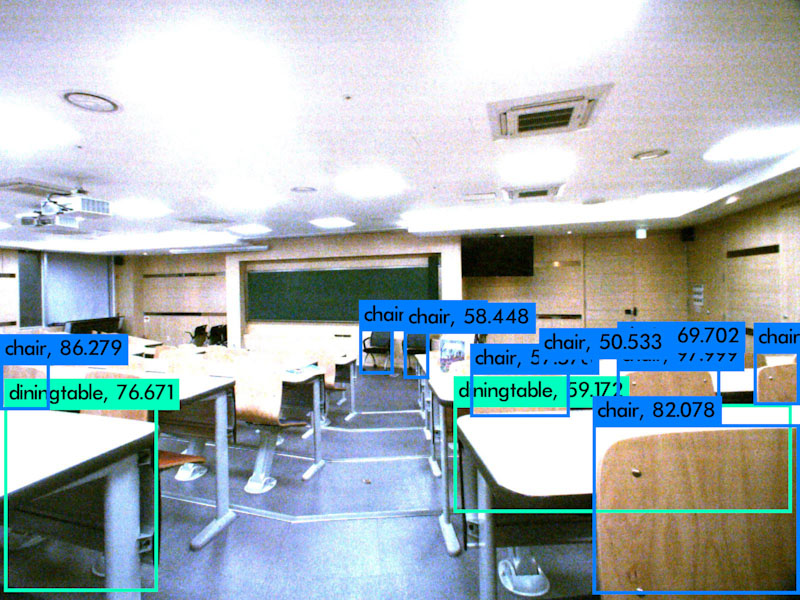} &
\includegraphics[width=0.195\linewidth]{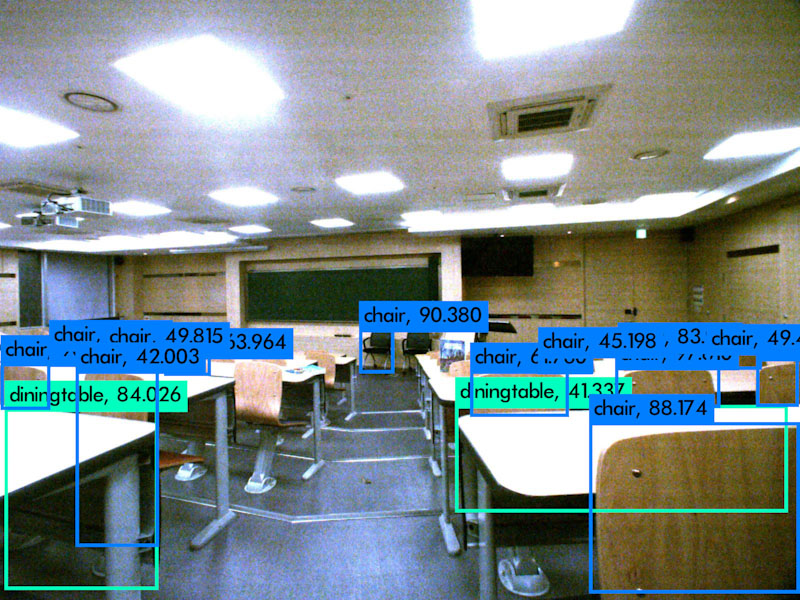} &
\includegraphics[width=0.195\linewidth]{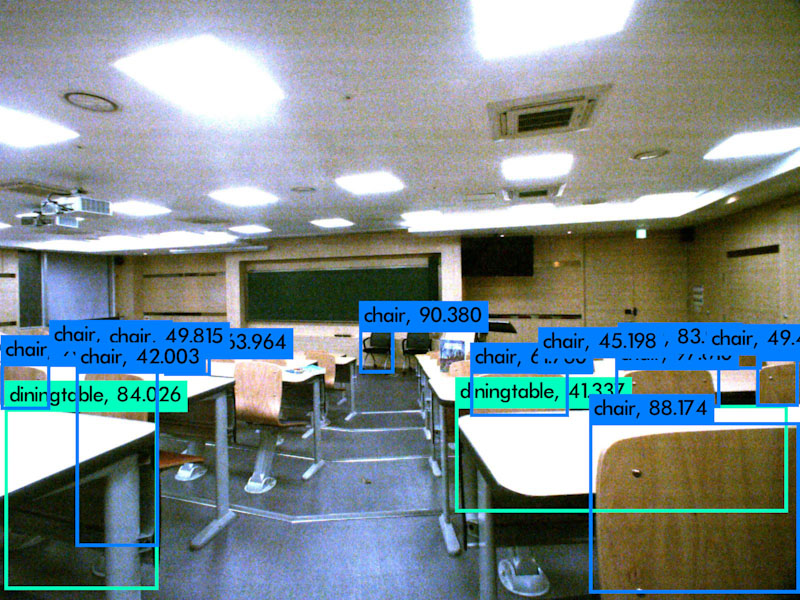} \\
{\footnotesize \textit{Ours} (0$dB$,67$ms$)} & {\footnotesize \textit{AE} (10$dB$,30$ms$)} & {\footnotesize \textit{Shim} (24$dB$,13$ms$)} & {\footnotesize \textit{Zhang} (24$dB$,13$ms$)} & {\footnotesize \textit{Kim} (24$dB$,10$ms$)} \\
\includegraphics[width=0.195\linewidth]{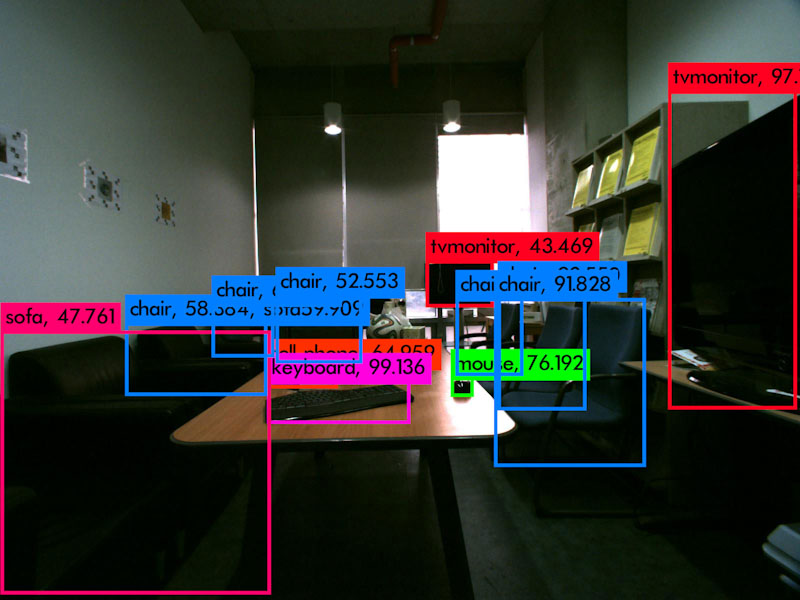} &
\includegraphics[width=0.195\linewidth]{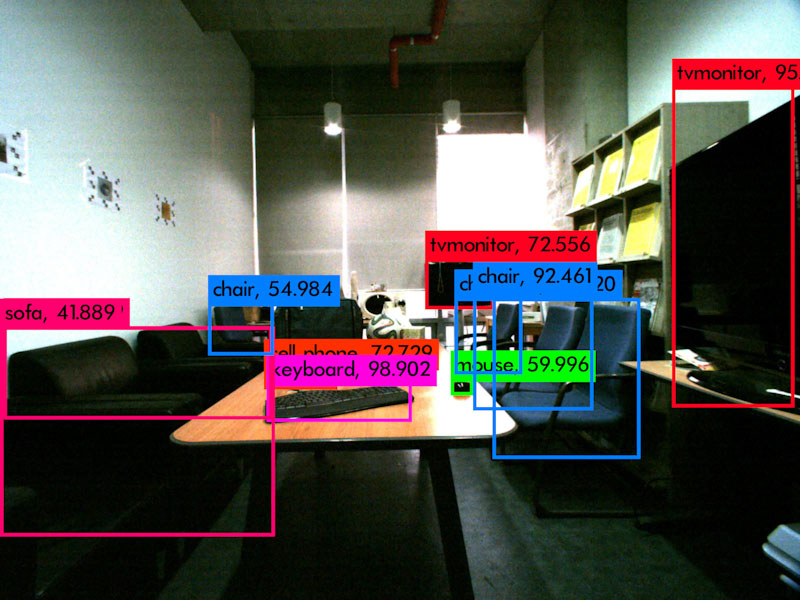} &
\includegraphics[width=0.195\linewidth]{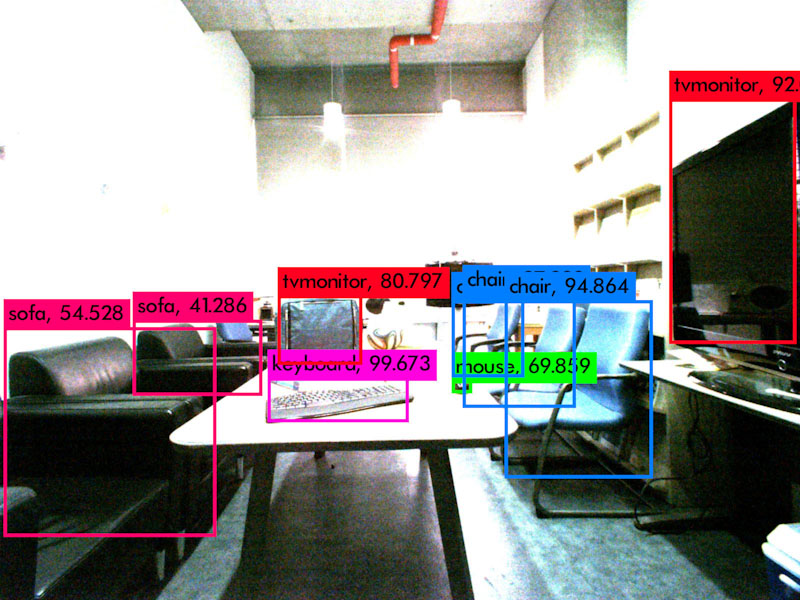} &
\includegraphics[width=0.195\linewidth]{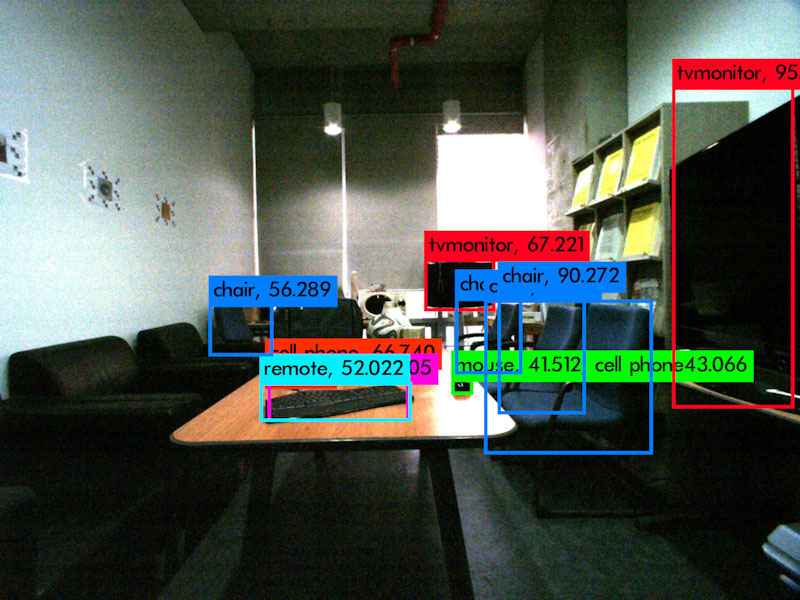} &
\includegraphics[width=0.195\linewidth]{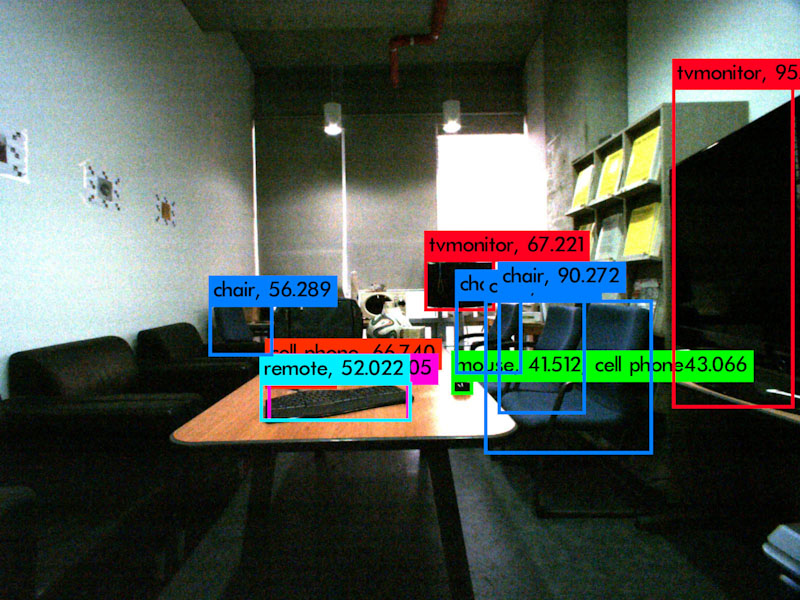} \\
{\footnotesize \textit{Ours} (8$dB$,55$ms$)} & {\footnotesize \textit{AE} (19.8$dB$,30$ms$)} & {\footnotesize \textit{Shim} (24$dB$,67$ms$)} & {\footnotesize \textit{Zhang} (24$dB$,16$ms$)} & {\footnotesize \textit{Kim} (24$dB$,16$ms$)} \\
\end{tabular}
\caption{{\bf Images selected by each algorithm and object detection results on \textit{Car4} dataset.} Except for the proposed method, other metrics suffer from several noise. 
Moreover, the high noise-level leads to misclassified objects and a low confidence value of detection. This problem clearly appears on Shim~\cite{shim2014auto}, Zhang~\cite{zhang2017active}, and Kim~\cite{kim2018exposure} results.}

\label{fig:detectionResult}
\vspace{-0.2in}
\end{figure*}

\subsection{Exposure Control Dataset}
\label{subsec:Dataset}

The real-time nature of AE algorithms makes their quantitative evaluation and comparison complex. In this paper, we provide a unique dataset developed specifically to compare such type of algorithms. For this purpose, we have constructed a stereo camera system with a $20cm$ baseline acquiring synchronized $1600 \times 1200$px images. Our dataset consists of 25 indoor/outdoor static scenes with various illumination levels and 13 manually labeled object classes\footnote{Person, Bicycle, Car, Firehydrant, Backpack, Sportsball, Chair, Mouse, Keyboard, Cellphone, Book, Scissors, and Tvmonitor.}. The outdoor scenes are captured by changing the exposure time from $0.1ms$ to $7.45ms$ with a step size $0.15ms$ and the gain from $0dB$ to $20dB$ with a step size $2dB$. For the indoor scenes, the exposure time and the gain range from $4ms$ to $67ms$ with a step size $3ms$ and from $0dB$ to $24dB$ with a step size $1dB$, respectively. Therefore, the dataset has 550 stereo image pairs for each scene.

We compare our algorithm with the camera built-in AE, Shim~\etal~\cite{shim2014auto}, Zhang~\etal~\cite{zhang2017active} and Kim~\etal~\cite{kim2018exposure}. Because none of these algorithms have been released publicly, we have re-implemented their image quality metric. After that, we select the best-exposed images from each metric in each scene of the whole dataset. These images are utilized for feature matching, pose estimation, and object detection experiments. \figfref{fig:noise_illustration} shows some of the well-exposed images selected by each metric. Generally on the entire dataset, camera built-in AE frequently captures under-exposed images in outdoor environment due to the sunlight, although it shows better performance under the indoor environment. Moreover, since the other metrics do not consider image noise into their metric, selected images are often saturated or highly noisy. In contrast, our metric selects images with rich textures and low noise compared to the existing approaches in both indoor and outdoor environments.

\subsection{Feature Matching and Pose Estimation}
\label{subsec:poseest}

\begin{table}[t]
\caption{\textbf{Quantitative comparison on feature matching and pose estimation.} (\textbf{\textcolor{red}{Red}} : Best, \textcolor{blue}{Blue} : Runner-up).}
\vspace{-0.2in}
\begin{center}
\resizebox{1\columnwidth}{!}{
\def\arraystretch{1.3}
\footnotesize
\begin{tabular}{|c|c||c|c|c||c|c|}
\hline
 & \textbf{Method} & $\mathbf{N_{feat}}$ & $\mathbf{\frac{N_{correct}}{N_{feat}}}$ & $\mathbf{\frac{N_{correct}}{N_{init}}}$ & $\mathbf{e_{r}}~(deg)$ & $\mathbf{e_{t}}~(m)$ \\
\hline
\hline
\multirow{5}{*}{\rotatebox{90}{\textbf{Indoor}}} & Ours & 13,099 & \textbf{\textcolor{red}{0.126}} & \textbf{\textcolor{red}{0.231}} & \textbf{\textcolor{red}{1.797}} & \textbf{\textcolor{red}{0.039}} \\
	& AE   & 15,080 & \textcolor{blue}{0.113} & \textcolor{blue}{0.215} & \textcolor{blue}{2.539} & \textcolor{blue}{0.067} \\
    & Shim~\cite{shim2014auto} & 45,488 & 0.058 & 0.117 & 6.199 & 0.179 \\
    & Zhang~\cite{zhang2017active} & \textcolor{blue}{46,354} & 0.043 & 0.089 & 5.938 & 0.263 \\
    & Kim~\cite{kim2018exposure} & \textbf{\textcolor{red}{46,483}} & 0.043 & 0.089 & 4.552 & 0.235 \\
\hline
\multirow{5}{*}{\rotatebox{90}{\textbf{Outdoor}}} & Ours & 42,026 & \textbf{\textcolor{red}{0.058}} & \textbf{\textcolor{red}{0.099}} & \textbf{\textcolor{red}{0.394}} & \textbf{\textcolor{red}{0.023}} \\
	& AE & 29,578 & 0.050 & 0.084 & 0.618 & 0.055 \\
	& Shim~\cite{shim2014auto} & \textbf{\textcolor{red}{46,322}} & \textcolor{blue}{0.052} & \textcolor{blue}{0.093} & \textcolor{blue}{0.470} & \textcolor{blue}{0.042} \\
	& Zhang~\cite{zhang2017active} & \textcolor{blue}{44,200} & 0.047 & 0.084 & 1.136 & 0.111 \\
	& Kim~\cite{kim2018exposure} & 40,427 & 0.049 & 0.087 & 1.132 & 0.137 \\
\hline
\end{tabular}
}
\end{center}
\label{table:matchingandpose}
\end{table} 

Camera pose estimation requires robust keypoints detection and matching which rely on strong and uniformly distributed gradients. Therefore, the feature matching ratio and the accuracy of the pose estimation reflect the performance of the camera exposure control. For this quantitative evaluation, we calibrate our stereo camera system using accurate calibration algorithms~\cite{zhang2000flexible,ha2017deltille}, then we get intrinsic, distortions, and extrinsic parameters of the two cameras admitting a re-projection error of $0.097$ px. We regard this result as a ground truth pose.

The initial feature matching is performed on undistorted images using ORB features~\cite{rublee2011orb} by brute-force matching. The 5-point algorithm~\cite{nister2004efficient} with Least median of squares (LMedS)~\cite{rousseeuw1984least} are adopted for a pose estimation. For the sake of repeatability, we intentionally avoided stochastic approaches like RANSAC for this estimation. The resulting rotation and translation error, $e_{r}$ and $e_{t}$, are computed as follows:

\vspace{-0.2in}
\begin{gather}
e_{r} = \arccos{\left(\left(\mathrm{Trace}\left(R_{GT} \cdot R^{T}\right) - 1\right) / 2\right)}, \label{eqn:rotationerr} \\
e_{t} = ||T_{GT} - T||_{2}, \label{eqn:transerr}
\end{gather}
where $R_{GT}, R, T_{GT}$ and $T$ denote the ground truth and estimated rotations and translations, respectively.

\Tabref{table:matchingandpose} contains the quantitative evaluation of our feature matching and pose estimation experiment. $N_{feat}$, $N_{init}$ and $N_{correct}$ denote the number of extracted local features, the number of initial matches, and the number of correct matches, respectively. We define a correct match as a match with very low reprojection error $(< 1e^{-4})$ calculated using ground truth intrinsic/extrinsic parameters. 

In every scenario, our algorithm demonstrates better keypoint repeatability since the percentage of inliers at every stage remains the highest compared to the other approaches. It should be noted that the number of extracted keypoints is not the representative of the image quality since high frequency noise is often triggering a large number of unwanted features. The quality of the pose estimation is also a good indicator confirming this assumption since our algorithm also ensures the highest accuracy. These results are coherent since we designed the proposed algorithm to specifically find an image with rich information and low noise level. In contrast, the other algorithms tend to overcompensate the low light condition (particularly in the indoor) with high gain, leading to images highly corrupted by noise.

\begin{table*}[t]
\caption{\textbf{Noise estimation performance and computation time comparisons.}}
\begin{center}
\def\arraystretch{1.2}
\footnotesize
\begin{tabular}{c|ccc|ccc|ccc|ccc}
\hline
\multirow{2}{*}{Method} & \multicolumn{3}{c|}{$\sigma = 1$} & \multicolumn{3}{c|}{$\sigma = 5$} & \multicolumn{3}{c|}{$\sigma = 10$} & \multicolumn{3}{c}{Time $(ms)$} \\ \cline{2-13}
 & $s$ & $b$ & MSE & $s$ & $b$ & MSE & $s$ & $b$ & MSE & min & mean & max \\ \hline \hline
Ours & 0.7817 & 0.7125 & 1.1186 & 0.4927 & 0.4475 & 0.4430 & 0.3729 & 0.3675 & 0.2741 & 62.003 & 88.305 & 91.485 \\
Immerkaer~\cite{immerkaer1996fast} & 1.6890 & 2.8040 & 10.7154 & 1.2738 & 1.5972 & 4.1736 & 0.9816  & 0.9875 & 1.9837 & 9.101 & 14.106 & 24.061 \\
Chen~\etal~\cite{chen2015efficient} & 0.3345 & 0.7024 & 0.5386 & 0.0623 & 0.0761 & 0.0097 & 0.0499 & 0.0563 & 0.0057 & 175.875 & 215.874 & 385.160 \\ \hline
\end{tabular}
\end{center}
\vspace{-0.2in}
\label{tab:noiseeval}
\end{table*}

\subsection{Object Detection Comparison}
\label{subsec:objdet}


We further evaluate the performance of each algorithm from the object detection point of view. The object detection results are obtained from MSCOCO~\cite{lin2014microsoft} pre-trained YOLOv3 object detector~\cite{redmon2018yolov3}. For this quantitative comparison, we compute Average Precision (AP) with two Intersection-over-Union (IoU) thresholds $AP_{50}$ and $AP_{75}$, which have been commonly used for the evaluation of object detection performance. Our algorithm gives the best object detection performance with both low and high IoU thresholds compared to the other methods: Ours($AP_{50} = 49.6$, $AP_{75} = 38.6$), AE($44.3$, $33.2$), Shim($43.5$, $23.6$), Zhang($37.8$, $25.8$), Kim($36.3$, $25.1$). \Figref{fig:detectionResult} contains the qualitative result. The images selected by our algorithm contain more successfully detected objects and with higher confidence. In fact, the confidence values of detection are higher than other techniques for the same objects thanks to the limited amount of noise in the image selected by our algorithm. In addition, our algorithm is more effective to detect small objects because they can be greatly contaminated by image noise~\cite{davies98detection} and an exposure condition.

\subsection{Noise-Based Metric Validation}
\label{subsec:validationnoisemetric}
To validate the performance of our noise-based metric $\sigma_{noise}$, we compare our method with Immerkaer~\cite{immerkaer1996fast} and Chen~\etal~\cite{chen2015efficient} on the TID2008 dataset~\cite{ponomarenko2009tid2008}, which has been widely used for the evaluation of noise estimation metrics. We follow the evaluation method based on the mean squared error (MSE) adopted by Chen~\etal~\cite{chen2015efficient} as follows:

\begin{equation}
MSE = E\left(\hat{\sigma} - \sigma\right)^{2} = b^{2}\left(\hat{\sigma}\right) + s^{2}\left(\hat{\sigma}\right)
\label{eqn:noisemse}
\end{equation}
where $\sigma$ and $\hat{\sigma}$ denote the ground truth and the estimated noise level and $s^{2}\left(\cdot\right)$ and $b\left(\cdot\right)$ denotes the variance and the bias of the estimator, respectively. Please refer to \cite{chen2015efficient} for further details. 

We add synthetic zero-mean white Gaussian noise with various variances values to each image, then estimate the noise level using each algorithm. \Tabref{tab:noiseeval} shows the noise estimation performance and computation time comparisons. Our metric is almost 10$\times$ more accurate compared to \cite{immerkaer1996fast} and 2.5$\times$ faster than \cite{chen2015efficient}. Therefore, our metric is the most suitable for real-time applications where both speed and reliable performance are essential. Note that our noise-based metric is re-implemented on MATLAB for a fair comparison in this experiment while our original code is in C++. The C++ results is available in the \secref{subsec:timeanalysis}.

\subsection{Ablation Study}
\label{subsec:Ablation Study}

\begin{figure}[t]
\small
\begin{tabular}{@{}c@{\hskip 0.005\linewidth}c@{\hskip 0.005\linewidth}c}
\includegraphics[width=0.32\linewidth]{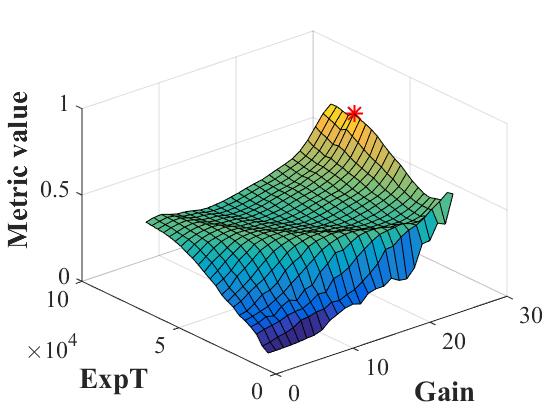} &
\includegraphics[width=0.32\linewidth]{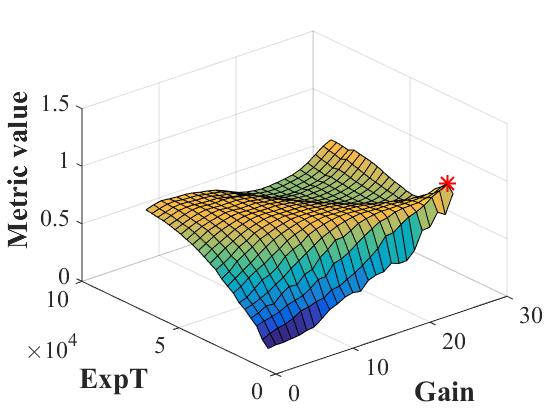} &
\includegraphics[width=0.32\linewidth]{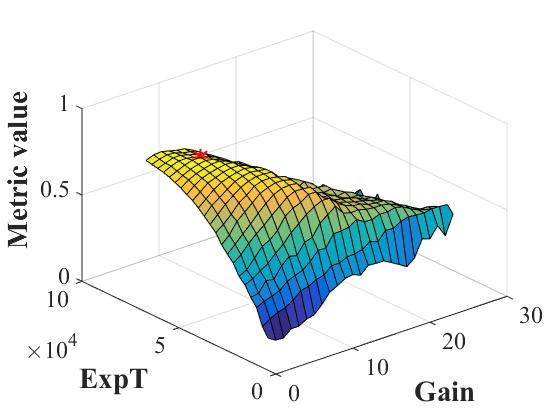} \\
($24dB$, $55ms$) & ($24dB$, $7ms$) & ($4dB$, $55ms$) \\
\includegraphics[width=0.32\linewidth]{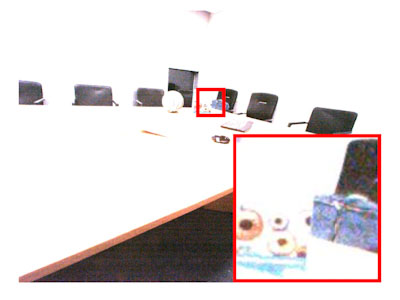} &
\includegraphics[width=0.32\linewidth]{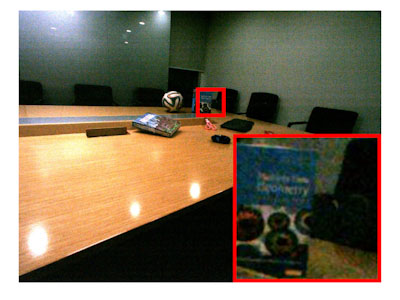} &
\includegraphics[width=0.32\linewidth]{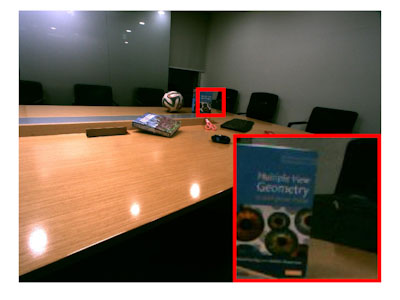} \\
(a) $L_{g}$ & (b) $L_{g}+L_{e}$ & (c) $L_{g}+L_{e}+\sigma_{noise}$ \\
\includegraphics[width=0.32\linewidth]{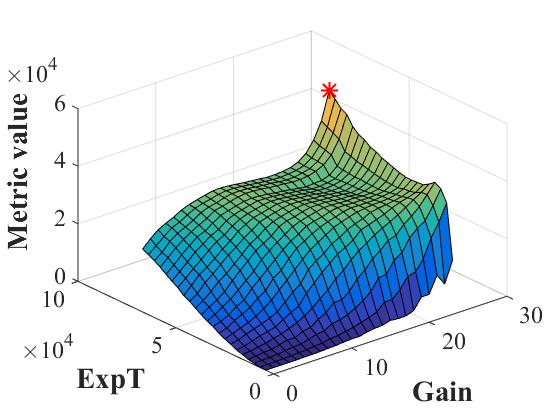} &
\includegraphics[width=0.32\linewidth]{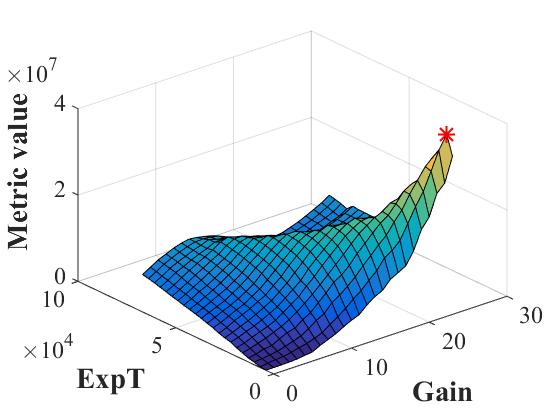} &
\includegraphics[width=0.32\linewidth]{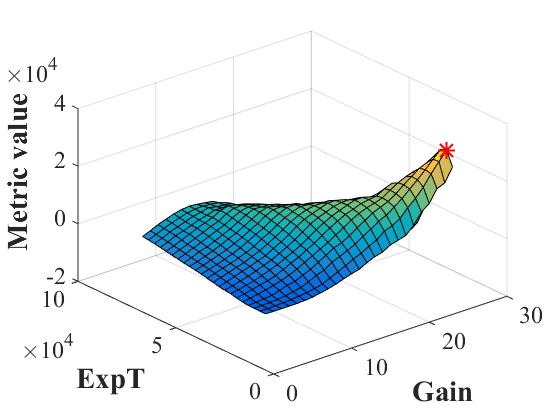} \\
($24dB$, $67ms$) & ($24dB$, $7ms$) & ($24dB$, $7ms$) \\
\includegraphics[width=0.32\linewidth]{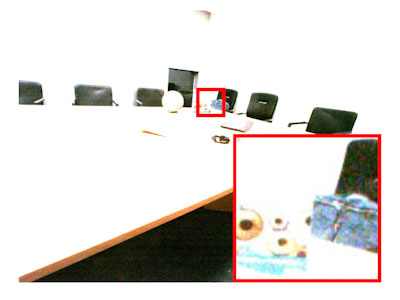} &
\includegraphics[width=0.32\linewidth]{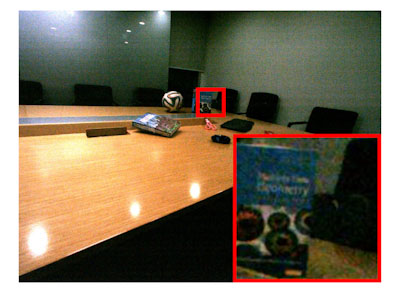} &
\includegraphics[width=0.32\linewidth]{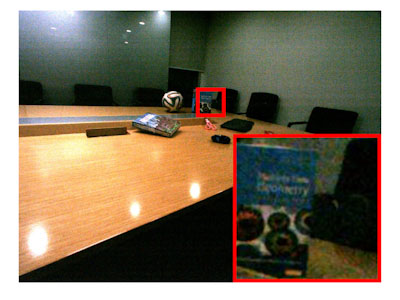} \\
(d) Shim~\cite{shim2014auto} & (e) Zhang~\cite{zhang2017active} & (f) Kim~\cite{kim2018exposure} \\
\end{tabular}
\caption{{\bf Performance comparison with various quality metrics.} Images in the second rows are selected by our algorithm with (a) $L_{gradient}$ only, (b) $L_{gradient} + L_{entropy}$ and (c) the proposed quality metric, respectively. (d)-(f) show results from the other algorithms for comparison. Each plot shows the quality metric surface with variable gain and exposure time, and the red star indicates the optimal (gain, exposure time) from each metric.}
\label{fig:ablationstudy}
\vspace{0.05in}
\end{figure}

In this section, we investigate the roles of gradient-, entropy- and noise-based metrics of our algorithm. In order to examine the behavior of various quality metrics, we plot the quality metric surface for each method in \figref{fig:ablationstudy}. The quality metric surface is convenient to visually analyze the convergence behavior provided by each algorithm.


Gradient-based metrics tend to get a high score at the high gain or long exposure time that make the image over-exposed and noisy. Moreover, they cannot distinguish noise and texture based on gradients only. This is clearly shown in the \figref{fig:ablationstudy} (a) and (d). Other algorithms considering this overexposure problem show different behaviors. These methods solve the exposure problem, but they are still suffering from several noise. \figref{fig:ablationstudy} (b), (e) and (f). In contrast to those approaches, the proposed noise-aware image quality metric (\eqnref{equ:ourmetric}) reduces noise problems significantly and the optimal point is not biased to both exposure time and gain. Therefore, image information are well preserved in the selected image with less noise (\figref{fig:ablationstudy} (c)).

\subsection{Exposure Control}
\label{subsec:expparamcontrol}

\begin{figure*}[t]
\footnotesize
\begin{tabular}{@{}c@{}c@{}c@{}c@{}c@{}c}
\includegraphics[width=0.164\linewidth]{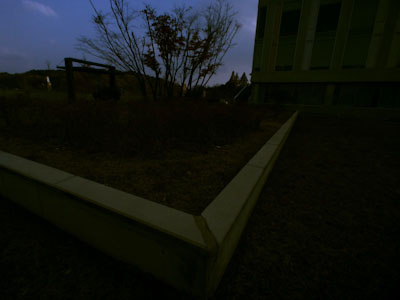} &
\includegraphics[width=0.164\linewidth]{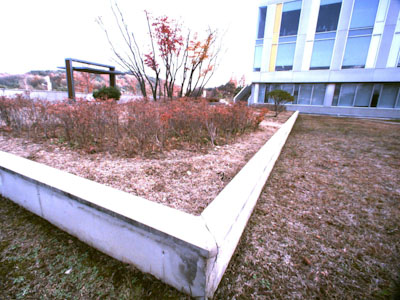} &
\includegraphics[width=0.164\linewidth]{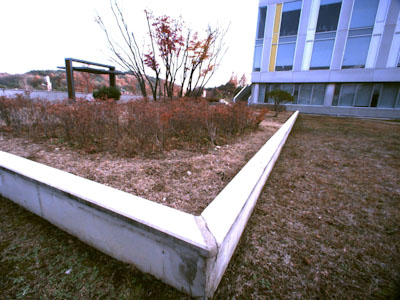} &
\includegraphics[width=0.164\linewidth]{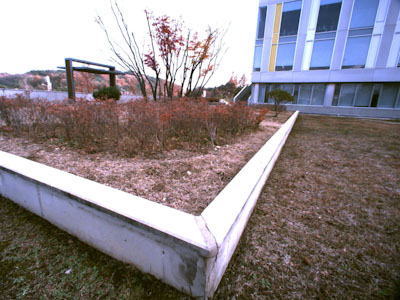} &
\includegraphics[width=0.164\linewidth]{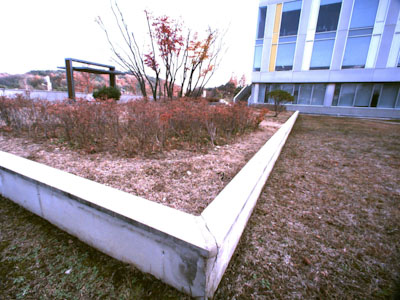} &
\includegraphics[width=0.164\linewidth]{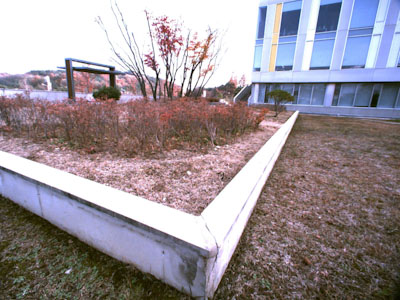} \\
Step 1 ($2.0, 1.10$) & Step 5 ($9.0, 4.95$) & Step 15 ($6.8, 5.66$) & Step 25 ($3.2, 7.48$) & Step 35 ($5.8, 6.78$) & Step 45 ($5.9, 6.83$)
\end{tabular}
\begin{tabular}{@{}c@{}c@{}c@{}c}
\footnotesize
\includegraphics[width=0.245\linewidth]{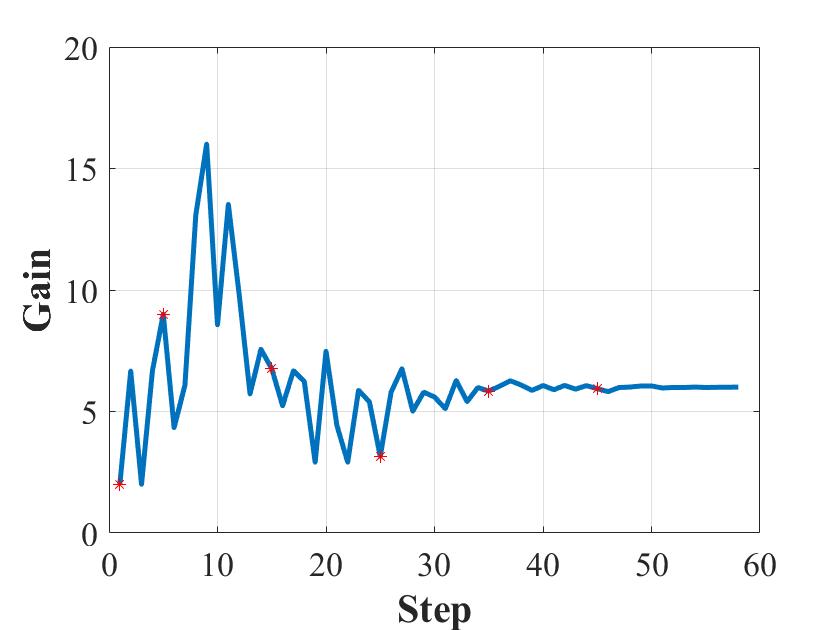} &
\includegraphics[width=0.245\linewidth]{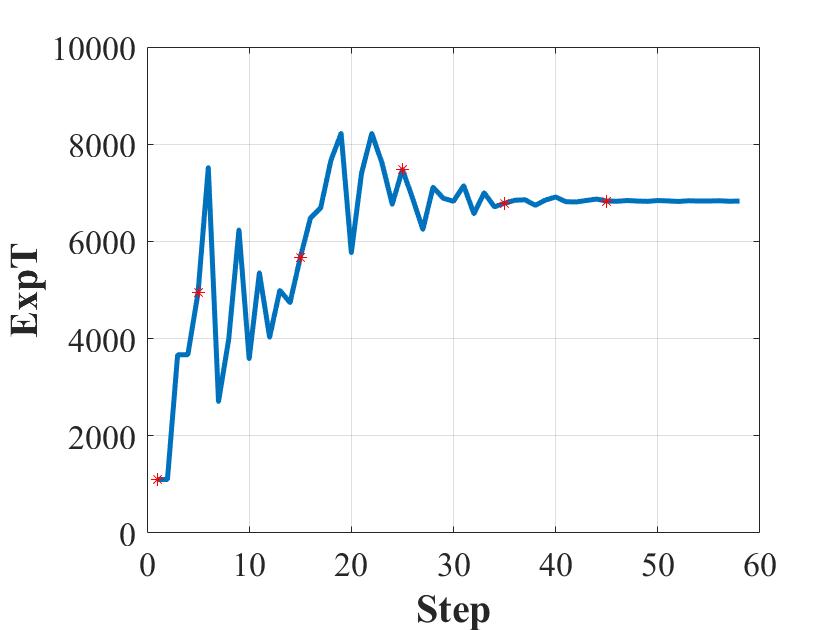} &
\includegraphics[width=0.245\linewidth]{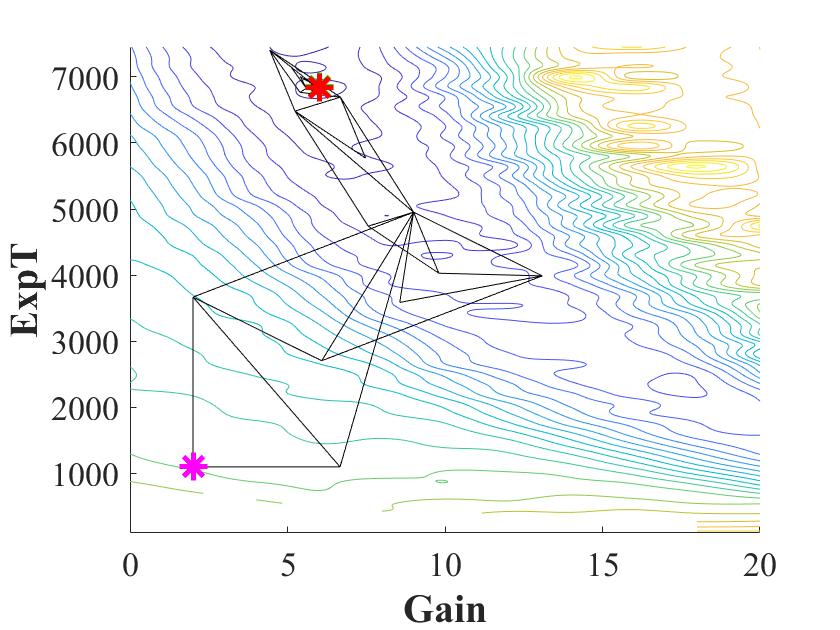} &
\includegraphics[width=0.245\linewidth]{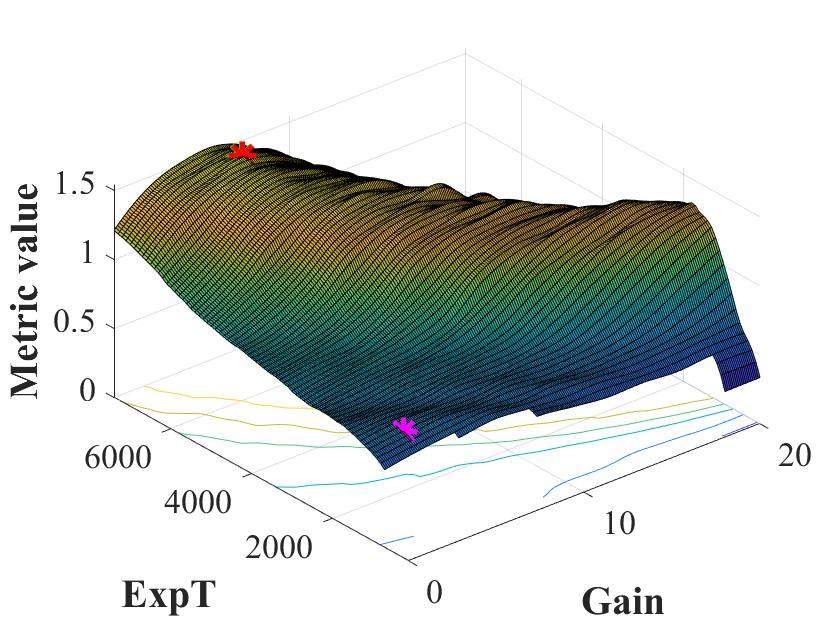} \\
Gain update & Exposure time update & Quality metric contour &  Quality metric surface
\end{tabular}
\vspace{0.05in}
\caption{{\bf Exposure parameter update sequence based on the gradient descent method.} The first row shows output images from each step and corresponding parameters ($dB$, $ms$). The second row shows the parameter convergence graphs, the quality metric contour and surface. Exposure time and gain are updated by the proposed algorithm~(\algref{alg:NMalgorithm}).}
\label{fig:gdcontrol}
\vspace{-0.2in}
\end{figure*}

In this section, we analyze the convergence speed and accuracy of the proposed control algorithm~(\algref{alg:NMalgorithm}). For the experiment, we first evaluate the proposed metric~\eqnref{equ:ourmetric} for 550 images of a static scene. Then, we estimate intermediate values with cubic interpolation with $0.1dB$ gain step and $1\mu s$ exposure time step to construct a quality metric surface. The proposed control algorithm starts from a random initial point on the contour, then we measure the number of steps until it converges. \Figref{fig:gdcontrol} shows intermediate output images, parameter update curves, the quality metric contour, and the quality metric surface. The parameters of intermediate output images are marked with red stars in parameter update curves. The initial, converged, and ground truth parameter are shown in the quality metric contour and surface with magenta, red, and green dots, respectively.

The proposed algorithm recursively updates the simplex by inspecting the solution space and find the optimal solution. As the simplex (black triangles in the quality metric contour of ~\figref{fig:gdcontrol}) is updated through the proposed control algorithm, it gets smaller quickly and converges to the ground truth solution. The converged parameters are ($5.9dB$, $6.83ms$), and the ground truth parameters are ($6dB$, $6.85ms$), respectively. The converged parameter and ground truth parameter are almost overlapped. Note that both gain and exposure time are almost converged after just 30 steps. This result demonstrates that the convergence of the proposed control algorithm is reliable and fast.

\subsection{Processing Time Analysis}
\label{subsec:timeanalysis}

\begin{table}[t]
\vspace{0.1in}
\caption{\textbf{Computation time analysis}}
\vspace{-0.2in}
\begin{center}
\resizebox{1\columnwidth}{!}{
\footnotesize
\def\arraystretch{1.1}
\begin{tabular}{lcc}
\hline
\multirow{2}{*}{\textbf{Method}} & \multicolumn{2}{c}{\textbf{Processing Time ($ms$)}} \\ \cline{2-3}
 & {$1600 \times 1200$ px} & {$800 \times 600$ px}  \\
\hline
\hline
     Gradient-based metric & $18.673$ & $3.232$\\
	 Entropy-based metric & $1.180$ & $0.316$\\
	 Noise-based metric & $88.868$ & $14.642$\\ \cline{1-3}
	 Total processing time & $108.721$ & $18.190$ \\
\hline
\end{tabular}
}
\end{center}
\label{table:timeanalysis}
\end{table} 

In order to determine whether the proposed algorithm is suitable for real-time applications, we have analyzed the computation times of each component. Our algorithm is implemented in C++ without multi-thread processing and tested on a i7-7700HQ@2.80GHz processor. Each process time is averaged over 1000 trials. We exclude the computation time of the NM method ($< 0.01ms$), which is negligible compared to the metric computation time.

\Tabref{table:timeanalysis} shows the computation times of our algorithm on $1600 \times 1200$px and $800 \times 600$px resolutions.  The proposed metric takes $108.72ms (9.2\mathrm{Hz})$ and $18.19ms (55.00\mathrm{Hz})$ for each resolutions. For the $800 \times 600$px resolution, our algorithm achieves real-time performance. The bottleneck of our algorithm is the noise-based metric calculation because it estimates noise levels in each channel, then averages them. To speed-up our algorithm, we can estimate the noise level using one channel solely (\textit{e.g.}, the green channel for the Bayer sensor). Moreover, it should be noted that the auto-exposure control algorithm does not require to run on high resolution images and can be computed at a smaller scale.

\section{Conclusion}
\label{sec:conclusion}

In this paper, we have proposed a noise-aware exposure control algorithm designed to capture well-exposed images. The proposed algorithm relies on a novel image quality metric coupling three complementary criteria based on the image gradient, entropy, and noise estimation. The synergy of these features demonstrates interesting properties to preserve sharp edges and rich textures while suppressing noise. Also, the proposed control algorithm guarantees fast and reliable exposure parameter convergence through simple and efficient searching strategy. Thanks to the light-weight computation and reliable convergence of the proposed algorithms, the proposed algorithm quickly and reliably produces desirable images that are suitable for various robotics and computer vision applications. In addition, we provided the exposure control dataset that consists of 25 indoor/outdoor scenes with 550 stereo images per scene and various camera exposure parameters. Our source code and dataset will be publicly available. In this paper, the proposed method did not fully prove their ability on dynamically changing environments, we will extend our proposed method to operate on dynamic environments robustly.

\bibliographystyle{./IEEEtran} 
\bibliography{./IEEEabrv,./IEEEexample}

\end{document}